\pdfoutput=1

\documentclass[11pt]{article}

\usepackage[final]{acl}

\usepackage{times}
\usepackage{latexsym}
\usepackage{booktabs}
\usepackage{multirow}
\usepackage{setspace}

\usepackage{subcaption} 

\usepackage[T1]{fontenc}

\usepackage[utf8]{inputenc}

\usepackage[final,nopatch=footnote]{microtype}

\usepackage{inconsolata}

\usepackage{graphicx}

\usepackage{soul} 
\definecolor{myyellow}{rgb}{1,1,0.6}
\definecolor{mygreen}{rgb}{0.6,1,0.6}
\definecolor{mypurple}{rgb}{0.85,0.75,0.9} 

\definecolor{myorange}{rgb}{1,0.75,0.4} 

\newcommand{\highlightA}[1]{%
  \begingroup
    \sethlcolor{myyellow}
    \hl{#1}
  \endgroup
}
\newcommand{\highlightB}[1]{%
  \begingroup%
    \sethlcolor{mygreen}%
    \hl{#1}%
  \endgroup%
}
\newcommand{\highlightC}[1]{%
  \begingroup%
    \sethlcolor{mypurple}%
    \hl{#1}%
  \endgroup%
}
\newcommand{\highlightD}[1]{%
  \begingroup%
    \sethlcolor{myorange}
    \hl{#1}%
  \endgroup%
}

\definecolor{darkblue}{rgb}{0.0,0.0,0.6}

\definecolor{plum}{rgb}{0.56, 0.27, 0.52}
\definecolor{pistachio}{rgb}{0.58, 0.77, 0.45}

\definecolor{chromeyellow}{rgb}{1.0, 0.65, 0.0}

\usepackage{amssymb}
\usepackage[svgnames]{xcolor} 
\usepackage{tcolorbox}
\tcbuselibrary{skins} 
\usepackage{enumitem}

\title{\textsc{Morables}: A Benchmark for Assessing\\ Abstract Moral Reasoning in LLMs with Fables}

\author{
  Matteo Marcuzzo $^{\spadesuit}$ \quad
  \textbf{Alessandro Zangari} $^{\spadesuit}$ \quad
  \textbf{Andrea Albarelli} $^{\spadesuit}$ \\
  \textbf{Jose Camacho-Collados} $^{\diamondsuit}$ \quad
  \textbf{Mohammad Taher Pilehvar} $^{\diamondsuit}$ \\
  $^{\spadesuit}$Dept of Environmental Sciences, Informatics and Statistics, Ca' Foscari University of Venice \\
  \normalsize{\texttt{\{name.surname\}@unive.it, albarelli@unive.it}} \\
  $^{\diamondsuit}$School of Computer Science and Informatics, Cardiff University \\
  \normalsize{\texttt{camachocolladosj@cardiff.ac.uk, pilehvarmt@cardiff.ac.uk}}
}

\newtcolorbox{contentbox}[1][]{ 
    colframe=black,            
    colback=white!0,          
    colbacktitle=gray!40,     
    arc=2mm,                  
    boxrule=1pt,              
    fonttitle=\bfseries\itshape\color{black},      
    center title,             
    enhanced,                 
    lower separated=true, 
    boxsep=2pt,               
    #1                        
}

\newtcolorbox{prompt}[1][]{ 
    colframe=black,            
    colback=white!0,          
    colbacktitle=gray!40,     
    arc=2mm,                  
    boxrule=1pt,              
    fonttitle=\bfseries\itshape\color{black},      
    center title,             
    enhanced,                 
    lower separated=false, 
    boxsep=2pt,               
    #1                        
}

\begin{document}
\maketitle
\begin{abstract}

As LLMs excel on standard reading comprehension benchmarks, attention is shifting toward evaluating their capacity for complex abstract reasoning and inference. 
Literature-based benchmarks, with their rich narrative and moral depth, provide a compelling framework for evaluating such deeper comprehension skills.
Here, we present \textsc{Morables}, a human-verified benchmark built from fables and short stories drawn from historical literature.
The main task is structured as multiple-choice questions targeting moral inference, with carefully crafted distractors that challenge models to go beyond shallow, extractive question answering. 
To further stress-test model robustness, we introduce adversarial variants designed to surface LLM vulnerabilities and shortcuts due to issues such as data contamination.
Our findings show that, while larger models outperform smaller ones, they remain susceptible to adversarial manipulation and often rely on superficial patterns rather than true moral reasoning. {This brittleness results in significant self-contradiction, with the best models refuting their own answers in roughly 20\% of cases depending on the framing of the moral choice.}
Interestingly, reasoning-enhanced models fail to bridge this gap, suggesting that scale -- not reasoning ability -- is the primary driver of performance.
\end{abstract}


\section{Introduction}
The evaluation of Large Language Models (LLMs) for natural language understanding remains a key challenge, driving the creation of benchmarks that reflect the depth and complexity of their evolving capabilities \cite{10.1145/3641289,BenchmarkingLLMs,dong-etal-2024-generalization,white2025livebench}. 
Traditional benchmarks such as GLUE \cite{wang-etal-2018-glue} and SuperGLUE \cite{superglue} have reached a point of performance saturation, and some of their tasks have been criticized for structural limitations.
A prime example is Natural Language Inference (NLI), a core reading comprehension task featured heavily in these benchmarks.
Despite its central role, studies have revealed significant shortcomings in how NLI datasets function as evaluation tools \cite{naik-etal-2018-stress,gururangan-etal-2018-annotation,nie-etal-2020-adversarial}.
For instance, \citet{mccoy-etal-2019-right} demonstrated that models trained and evaluated on MNLI \cite{williams-etal-2018-broad}
often rely on superficial heuristics rather than genuine understanding. 
While adversarial datasets have been introduced to mitigate the influence of spurious cues, they often lead to artificial examples that may not reflect real-world language use \cite{Bihani2025}. 

Recognizing these limitations, recent research has begun to explore alternative benchmarking approaches that better capture the richness of human language understanding \cite{ghosh-srivastava-2022-epic,sravanthi-etal-2024-pub,li2024when}. One promising direction involves literature-based benchmarks, which are designed to probe higher-order reasoning and comprehension using authentic narratives. 
Unlike many traditional tasks, benchmarks derived from literary texts present models with more natural linguistic structures and challenge them to interpret implicit themes that require deeper inferential skills \cite{kocisky-etal-2018-narrativeqa}. 

\begin{figure}[t]
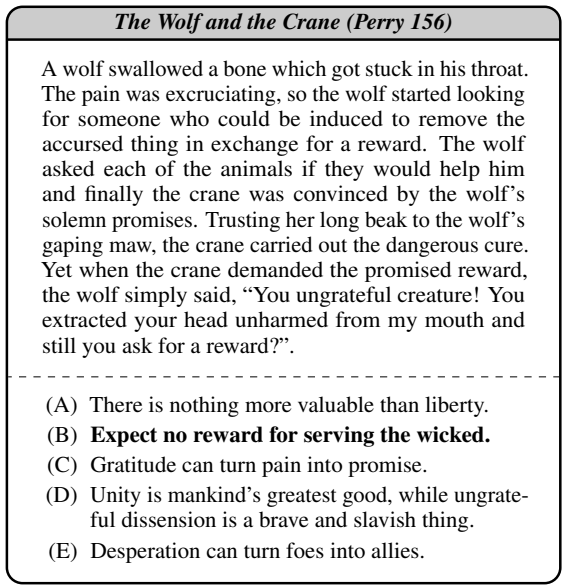
 
    \footnotesize
    \centering
    \scalebox{0.95}{\begin{contentbox}[title={The Wolf and the Crane (Perry 156)}]

        A wolf swallowed a bone which got stuck in his throat. The pain was excruciating, so the wolf started looking for someone who could be induced to remove the accursed thing in exchange for a reward. The wolf asked each of the animals if they would help him and finally the crane was convinced by the wolf's solemn promises. Trusting her long beak to the wolf's gaping maw, the crane carried out the dangerous cure. Yet when the crane demanded the promised reward, the wolf simply said, ``You ungrateful creature! You extracted your head unharmed from my mouth and still you ask for a reward?''.
        
        \tcblower
         
        \begin{enumerate}[label=(\Alph*),  itemsep=2pt, parsep=0pt, left=0pt] 
            \item There is nothing more valuable than liberty.
            \item \textbf{Expect no reward for serving the wicked.}
            \item Gratitude can turn pain into promise.
            \item Unity is mankind's greatest good, while ungrateful dissension is a brave and slavish thing.
            \item Desperation can turn foes into allies.

        \end{enumerate}
    \end{contentbox}}

    \caption{A sample entry from the \textsc{Morables} core MCQA dataset, with the correct option (B) in bold.}
    \label{fig:example}
\end{figure}

Building on this line of work, we propose the use of moral fables drawn from literary tradition as a novel tool to assess moral inference in modern LLMs. 
We introduce \textsc{Morables}, a curated benchmark of 709 short stories and fables primarily drawn from Western literary tradition. 
Each entry includes a high-quality transcription or translation of an original fable, paired with a moral attributed to the original author or translator. 
Our proposal consists of a human-verified Multiple Choice Question Answering (MCQA) task where the original moral is presented alongside a diverse set of four alternative options
(Figure \ref{fig:example} provides an example). 
We create multiple variants of our benchmark based on story and choice modifications to test for potential biases, including data contamination and shallow shortcut answering. Lastly, alongside MCQA evaluation, we assess the quality of LLM-generated free-text morals via human annotation.



%

Our findings show that, although larger models perform well, {certain} adversarial modifications and alternative evaluation procedures suggest that these models may rely more on {thematic matching with story events}
than genuine moral reasoning. {For instance, this weakness is illustrated by our discovery that even the best models contradict their own judgments in roughly 20\% of cases when the moral inference task is reframed.} Furthermore, our analysis of reasoning-augmented models indicates that explicit reasoning contributes less to performance than overall model size. Lastly, our experiments with human evaluation of free-text morals indicate that this task is highly challenging and potentially open ended, with semantic similarity to the original moral showing only a weak correspondence. This highlights moral inference as a complex and multi-faceted task.

\section{Related Work}

\subsection{Fables, Stories and Morals}

\paragraph{Moral Fables in NLP}
The only previous work explicitly addressing fables and morals in NLP is by \citet{guan-etal-2022-corpus}, who introduced a bilingual dataset of crowd-sourced stories and morals.
However, their broader definition of a story -- ``a series of coherent events involving several interrelated characters, implying support or opposition of some behavior'' -- results in many entries that deviate from classical fables, including Wikipedia excerpts and forum anecdotes. 
Our contribution differs substantially, focusing solely on historically sourced fables which are manually verified. 

\paragraph{Narrative and story datasets}
Several datasets have been proposed to investigate machine reasoning within the context of stories, specifically targeting the tasks of selecting and generating suitable story endings \cite{mostafazadeh-etal-2016-corpus,10.1609/aaai.v33i01.33016473}.
Notable examples include ROCStories \cite{mostafazadeh-etal-2016-corpus}, WritingPrompts \cite{fan-etal-2018-hierarchical},
roleplayerguild \cite{louis-sutton-2018-deep}, 
and PG-19 \cite{Rae2020Compressive}, but none are explicitly centered on moral narratives.

\paragraph{Morality and ethics datasets}
Previous work explored morality and ethics, though in ways fundamentally different from fable-based morals, primarily aiming to evaluate machine alignment with human ethical reasoning. Notable 
datasets include
Moral Stories \cite{emelin-etal-2021-moral}, ETHICS \cite{hendrycks2021aligning}, and Scruples \cite{Lourie_LeBras_Choi_2021}, each focusing on scenarios requiring morally appropriate actions or ethical judgments.



\subsection{Machine Reading Comprehension}

Most existing machine reading comprehension tasks are structured as question answering, with broad categorizations based on the types of questions and answers \cite{qiu2019surveyneuralmachinereading,rogers-rumshisky-2020-guide,QAExplosion}.
\textit{Span-based} 
tasks involve selecting a continuous span of text from a given context as the answer.
Notable datasets in this category include SQuAD/SQuAD2.0 \cite{rajpurkar-etal-2016-squad,rajpurkar-etal-2018-know}, TriviaQA \cite{joshi-etal-2017-triviaqa}, MS Marco \cite{nguyen2017ms}, NewsQA \cite{trischler-etal-2017-newsqa}, and HotpotQA \cite{yang-etal-2018-hotpotqa}, covering a wide range of domains and settings. 
\textit{Cloze-style} tasks, on the other hand, involve filling in blanks within a sentence or passage, as seen in datasets like CNN/Daily Mail \cite{NIPS2015_afdec700} and WikiReading \cite{hewlett-etal-2016-wikireading}. 

The most adopted format is the \textit{MCQA} task, which offers a pre-defined set of options to choose from as answers to a passage. 
These choices are typically created by domain experts and are often designed to sway the respondent's decision. 
Prominent datasets include RACE \cite{lai-etal-2017-race}, MMLU \cite{hendrycks2021measuring}, ARC \cite{ARC}, and MMLU-pro \cite{wang2024mmlupro}, all of which are widely used in LLM evaluation. However, it is worth noting that this approach is vulnerable to annotation artifacts and shallow cues, leading models to perform well without true understanding \cite{rogers-rumshisky-2020-guide}. 

Lastly, \textit{free-form answer} tasks have also been explored in datasets like NarrativeQA \cite{kocisky-etal-2018-narrativeqa} and DuReader \cite{he-etal-2018-dureader}. 
While metrics such as BLEU \cite{papineni-etal-2002-bleu} and, more recently, BERTScore \cite{BERTScore} have been used for evaluation, they still show limited alignment with human judgments of similarity \cite{leung-etal-2022-semantic,herbold2024semantic}.

\section{The \textsc{Morables} Benchmark}
\label{sec:benchmark}

In this section, we describe the \textsc{Morables} benchmark, which is publicly available.\footnote{\url{https://huggingface.co/datasets/cardiffnlp/Morables}} 

\subsection{Data Collection}

\textsc{Morables} is sourced from a variety of open sources, detailed in Appendix~\ref{sec:appendix_dataset}. Aesop, the most prominent Western fabulist, serves as the primary source and has influenced many later authors, such as La Fontaine, to compose fables in a similar style.

Our definition of fable follows that of \citet{Jose01022005}, which defines it as characterized by three main aspects: \textit{(i)} they are short, \textit{(ii)} they feature talking animals with a metaphorical meaning (though some fables contain historical figures),
and \textit{(iii)} they involve morally significant actions and outcomes.


\paragraph{Collection procedure}

The data collection was carried out in the following steps. First, we identified suitable sources, defined as those that \textit{(i)} contain both a fable and an associated moral, \textit{(ii)} are derived from historical literature and official translations, and \textit{(iii)} are in the open domain. 
Then, the stories and their morals were extracted from the relevant websites or books. Since multiple sources may report the same story, we conducted an in-depth duplicate removal process. This involves the utilization of similarity scores such as word Intersection over Union (IoU) and BERTScore \cite{BERTScore}. For Aesop's fables, its Perry Index \cite{Perry1952} is also used to check for duplicates.  

At the end of this process, we obtained 709 pairs of fables and their corresponding morals, {whose statistics are reported in 
Table~\ref{tab:dataset_statistics}.
The fables are primarily short texts, typically composed of a few long sentences that detail conversations between characters. In contrast, the morals are expressed as single, concise statements.
A more detailed analysis of the textual content, including common themes, characters, and readability, 
can be found in Appendix \ref{sec:appendix_dataset}.

\begin{table}
    \centering
        \small
        \setlength{\tabcolsep}{12pt}
    \begin{tabular}{ll}
    \toprule
       \textbf{Text Statistic} & \textbf{Value} \\
    \midrule
        Number of fable/moral pairs
            & 709 
                 pairs \\
        Avg length: fables 
            & 133.4 
                 words \\
        Avg length: morals
            & 11.6 
                 words \\
        Unique words: fables + morals
            & 7,278 
                 words \\
        Avg. number of sentences: fables 
            & 5.6
                 sentences \\
        Avg. sentence length: fables 
            & 25.0
                 words \\
            \bottomrule
    \end{tabular}
       \caption{{Statistics for the original fables and morals in \textsc{Morables}. Word and sentence counts are calculated using the \texttt{word\_tokenize} and \texttt{sent\_tokenize} functions from \texttt{NLTK} \cite{bird-2006-nltk}.}}
    \label{tab:dataset_statistics}
\end{table}


\subsection{Core Dataset Construction}
We develop a MCQA-based language understanding task based on the retrieved fables and their morals. The \textbf{\textsc{Morables}} dataset features carefully constructed answer choices, each designed to test the respondent's decision-making process.
\paragraph{MCQA distractors} 

The \textsc{Morables} core dataset requires models to select the correct moral among five candidates. 
To create a challenging benchmark, we focus on two key objectives: \textit{(i)} developing plausible negative alternatives,
thereby reducing the effectiveness of superficial cues and shortcut learning, and \textit{(ii)} incorporating human validation to ensure both dataset quality and that distractors are not overly plausible or misleading.

In the following paragraphs, we outline the procedure for generating challenging negative choices. We implement a systematic information extraction process to identify similar characters and entities, salient features and alternatives, and to incorporate distractor adjectives and newly generated morals. All extraction processes were performed using GPT-4o (prompts detailed in Appendix \ref{sec:appendix_prompts}), with outputs verified by human annotators for accuracy and appropriateness.

\paragraph{(1) Similar-character moral}
Our first distractor choice consists of a moral from another story in the dataset, featuring similar characters but with a different development.
This approach aims to reveal potential biases or shortcut strategies based solely on the entities in the story. 
However, not all fables contain characters that appear in other stories. To address this, we expand our search by generating plausible alternatives for each character (\textit{e.g.}, frogs $\rightarrow$ toads, fox $\rightarrow$ jackal) and matching stories based on these substituted entities.

\paragraph{(2) Trait-injected moral} 
\label{par:info_extract}
Our second distractor is also derived from the moral of a different fable, but it is modified to further assess the models' reliance on superficial cues. Specifically, we extract prominent features or traits of the fable's characters and insert them into an incorrect moral. As with the previous option, we prioritize stories with similar characters to make the choice more compelling, while ensuring that the moral selected is distinctly different from the first option. 

\paragraph{(3) Feature-based moral}
Our third distractor is an LLM-generated moral, where the model is given only the 
traits and characteristics of the fable's characters and must base the moral solely on those attributes. Although this moral may be structurally similar to authentic ones, it is necessarily detached from the narrative's events or overarching message, as it lacks access to the story's context. 

\paragraph{(4) Partial-story moral}
The final distractor is also LLM-generated, but it is based on an excerpt from the story (specifically, the first 10\% of sentences). This option is particularly challenging to craft, as the moral or the source story may sometimes be inferred from the excerpt, especially in shorter fables. 
To mitigate this risk, we structure the generation prompt to encourage creativity and a unique moral, providing a hand-crafted example as guidance.

\paragraph{Proof-checking}
The most important factor when creating negative answer choices is that they remain plausible but fundamentally incorrect. While some choices may be partially correct or overly generic, the original moral must always be the correct answer. To ensure this, we conduct both a semi-automatic similarity check and a human-driven validation of our benchmark. 
Each distractor is assessed for its similarity to the original one using standard similarity measures (word IoU and BERTScore). Morals that exhibit high similarity ($>0.5$ IoU, $>0.4$ BERTScore F1) are then manually evaluated -- roughly 15\% of feature-based morals and 25\% of partial-story morals. The main guideline for evaluation is to accept a generated moral if it is coherent but fails to correctly capture the underlying meaning of a fable. 
We further conduct an in-depth human evaluation of both the dataset and the generated options, detailed in the next Section. When a suitable alternative could not be AI-generated, it was curated manually.

Finally, we examined additional options for the dataset, including the incorporation of opposite morals, as detailed in Appendix \ref{sec:appendixBinary}. The appendix presents related experiments and our rationale for excluding these options from the final benchmark.

\subsection{Human Annotation and Validation}
\label{sec:human_ann}

It is essential to acknowledge that human oversight or
annotator bias in dataset creation can result in overly challenging questions, which may lead to distorted metrics \cite{Rubin16112024}.
In our study, this may occur due to: \textit{(i)} alternative options appearing more suitable for the fable, \textit{(ii)} options with meanings that are too similar, or \textit{(iii)} the original moral reflecting outdated ethical values, making it less appealing. To address these issues, we implement a comprehensive human validation procedure with multiple annotators from diverse backgrounds. Annotation is conducted using Label Studio \cite{LabelStudio}, with a screenshot of the GUI provided in Appendix~\ref{app:user_study}.

\paragraph{Human validation procedure}
Five graduate-level English native speakers from diverse academic backgrounds (Computer Science, Neuroscience, Theology, and Literature) were selected as annotators after a
screening 
where they annotated 20 fables of varying difficulty, from straightforward to intentionally ambiguous. Cohen's Kappa scores \cite{McHugh2012-br} indicated substantial agreement (0.6–0.75). However, some difficult fables proved ambiguous even for 
the
annotators, underscoring the 
need for
human-driven benchmark refinement.

The dataset was divided into batches, with 
each fable receiving 
two independent annotations.
Annotators were presented with each fable and five candidate morals, and were asked to select the most appropriate. To identify 
ambiguity
we allowed annotators to choose multiple answers, instructing them to do so only if they found the options equally valid. Fables with two incorrect annotator answers were flagged as ambiguous and manually reviewed, generating alternative answers when appropriate. In total, 152 fables contained at least a moral or distractor option that needed to be replaced. 
Morals reflecting outdated values were replaced with updated versions coherent with the story, with new distractors created as needed.

\subsection{Dataset Variants: TF and NOTO}
\label{sec:tf_noto}
Following the work of \citet{salido2025othersgeneraltechniquedistinguish} and \citet{wang-etal-2025-llms-may}, we also explore two alternative evaluation procedures: \textit{(i)} substituting all the correct answers with ``None of the other options'' (\texttt{NOTO} version); and \textit{(ii)} transforming the task into a binary one, where each choice is framed as a True or False question responding to the prompt ``\textit{True or False: The moral is ... }'' (\texttt{TF} version).
We also conduct a short investigation into the impact of answer ID naming and token bias \cite{wei-etal-2024-unveiling}, detailed in Appendix~\ref{sec:appendix_output_tokens}. Briefly, answer IDs had little impact on results, but token bias appeared in some open-weight models.

\subsection{Adversarial MCQA}
LLMs are trained on vast amounts of data, 
which likely includes classical fables such as those found in this dataset. Although some lesser-known fables may not have appeared during pre-training, contemporary research must assume that models have seen the majority of internet-available content. 
To address this, we introduce a set of adversarial modifications to the \textsc{Morables} dataset, designated as the \texttt{ADV} variant.
Here, LLMs are challenged through subtle alterations to the fables or multiple-choice options, detailed below. Figure \ref{fig:fable_modifications} illustrates an example of all possible fable modifications.

\begin{figure}[t]
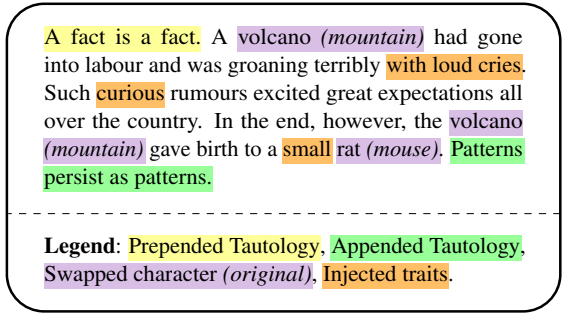

    \footnotesize
    \centering
    \scalebox{0.95}{
    \begin{tcolorbox}[
        colframe=black,            
        colback=white!0,          
        arc=5mm,                    
        boxrule=1pt,                
        enhanced,                   
    ]
    
   \highlightA{A fact is a fact.} A \highlightC{volcano \textit{(mountain)}} had gone into labour and was groaning terribly \highlightD{with {loud} cries}. Such \highlightD{curious} rumours excited great expectations all over the country. In the end, however, the \highlightC{volcano \textit{(mountain)}} gave birth to a \highlightD{small} \highlightC{rat \textit{(mouse)}}. \highlightB{Patterns persist as patterns.}
    
        \tcblower
        \textbf{Legend}:
\highlightA{Prepended Tautology}, 
\highlightB{Appended Tautology}, 
\highlightC{Swapped character \textit{(original)}},
\highlightD{Injected traits}.

    \end{tcolorbox}
    }
    \caption{An example with multiple adversarial modifications; colors indicate alteration types. Original characters (in parentheses) are not present in the actual text.}
    \label{fig:fable_modifications}
\end{figure}

\paragraph{(1) Character swap} 
The first modification swaps characters and entities within the fable with plausible alternatives.
This approach tests whether models are simply associating specific characters with morals. 
The modification is implemented using basic string matching and replacement.

\paragraph{(2) Trait injection} 
The injection of major traits and characteristics in fables follows the same process as trait-injected morals described in Section \ref{par:info_extract}.
We first select a different fable-moral pair and, when appropriate, inject character traits from the candidate fable into the story being modified.
Not all adjectives are used, as some may cause incoherence.
As before, we prioritize fables with similar characters, excluding those already used. The candidate pair's moral is then included among negative answer choices.

\paragraph{(3) Tautology injection}
Our final modification prepends or appends short, self-contained sentences -- specifically, tautologies (\textit{e.g.}, ``It is what it is'') -- designed to add no additional meaning to the story. We begin with a manually compiled list of tautology-moral pairs, used as a seed to automatically generate a larger set (with some detailed in Appendix \ref{sec:appendix_dataset}). Each tautology is paired with a moral matching its message (\textit{e.g.}, ``Accept things as they are''). The list is then manually filtered to exclude sentences that could alter the story's meaning. 

\paragraph{Proof-checking}
Since these modifications introduce new morals, we verify that they are not semantically similar to the correct choice using the same procedure described earlier. We test both adding and substituting the new moral among the answer choices (limiting the total to five), finding that the number of choices does not significantly affect model performance (see Appendix \ref{sec:appendix_output_tokens}).
\subsection{Summary of Dataset Variants}

Table~\ref{tab:benchmark_statistics} summarizes the composition and scale of each variant of the \textsc{Morables} benchmark. 
The \texttt{TF} variant reframes the dataset into 3,545 distinct True/False statements (709 fables × 5 choices). Whether and how the distractors are modified in the \texttt{ADV} variant depends on the combination of modifications being tested (see Section \ref{subsec:advmcqa}).

\begin{table}[t!]
    \centering
    \scalebox{0.88}{

    \begin{tabular}{l l l}
    \toprule
       \textbf{Variant} & \textbf{\# of entries} & \textbf{Choice answers} \\
    \midrule
        Core
            & 709
                & GT + distractors \\
        \texttt{NOTO}
            & 709 
                & NOTO + distractors \\
        \texttt{TF}
            & 3,545
                & GT or distractor  \\
        \texttt{ADV}
            & 709 
                & GT + (adversarial) distractors \\
            \bottomrule
    \end{tabular}
    }
    \caption{Scale and answer choice composition for the four \textsc{Morables} dataset variants.}
    \label{tab:benchmark_statistics}
\end{table}
\section{Evaluations}

We follow standard LLM evaluation for MCQA tasks \cite{zheng2024large}, prompting models to predict the answer ID token (\textit{e.g.}, A, B, C). The first generated token, stripped of spaces and newlines, is used as the answer. For models with an accessible temperature parameter, we set it to 0 for deterministic results. While output probabilities for option tokens could be used in open models, we confirm this yields equivalent results
(see Appendix~\ref{sec:appendix_output_tokens}). Choice options are shuffled to give each token roughly equal chance of being correct.

\paragraph{LLMs evaluated}

We report results for several recent LLMs, 
{grouped for clarity as: \textit{(i)} small-open, \textit{(ii)} large-open, \textit{(iii)} large-closed, and \textit{(iv)} reasoning.}
Among open models, we test Llama 3.3 (70B) \cite{grattafiori2024llama3herdmodels}, Mixtral 8x22B \cite{mistral2024mixtral8x22b}, and DeepSeek V3 (2024-12-26) \cite{deepseekai2024deepseekv3technicalreport} as large-model representatives. To assess smaller models, we include Llama 3.1 (8B), Mistral 7B \cite{jiang2023mistral7b}, and Qwen 2.5 (7B) \cite{qwen2025qwen25technicalreport}. For closed-source models, we select GPT-4o (2024-08-06) \cite{openai2024gpt4ocard}, Gemini 2.0 Flash (2025-02-05) \cite{geminiteam2025geminifamilyhighlycapable} and Claude 3.5 Sonnet (2024-10-22) \cite{anthropic2024claude35sonnet}, owing to their strong performance and stability.
We also evaluate two reasoning models: the closed-source GPT-o3-mini (2025-01-31) \cite{openai2024gpto3mini} and the open-weight DeepSeek R1 (2025-01-20) \cite{deepseekai2025deepseekr1incentivizingreasoningcapability}. 
While more advanced models such as OpenAI's o1 exist, their cost is prohibitive even for a benchmark of this size.

We present results using the best prompt identified in our preliminary tests, which provides simple guidelines and a single example (one-shot). Initial experiments showed that one example significantly improved performance, while additional examples had minimal effect. {These results may thus be considered an upper bound, as zero-shot settings yield inferior performance (4--6\% drop in accuracy).}
A comparison of zero- and one-shot results, along with all prompts used, is provided in Appendix \ref{sec:appendix_prompts}.

\subsection{Results}


\begin{table}
  \centering
  \setlength{\tabcolsep}{7pt}
  \scalebox{0.88}{
  \begin{tabular}{llc}
    \toprule
     & \textbf{Model}           
            & \textbf{Accuracy \%} \\
    \midrule

    \multirow{3}{*}{\centering\textit{Small-open}}
    &  Mistral 7b
            & 28.4 \scriptsize{$\pm$ 0.8}  \\
     & Llama 3.1 8b
            & 34.0 \scriptsize{$\pm$ 1.5}  \\

    &  Qwen 2.5 7b
            & 46.8 \scriptsize{$\pm$ 0.3}  \\

    \midrule
    \multirow{3}{*}{\centering\textit{Large-open}}
    &  Mixtral-8x22b
            & 56.9 \scriptsize{$\pm$ 1.4}  \\
    &  Llama 3.3 70B
            & 73.6 \scriptsize{$\pm$ 0.6}  \\
    &  DeepSeek V3
        & 70.6  \scriptsize{$\pm$ 0.8} \\
    \midrule
    \multirow{3}{*}{\centering\textit{Large-closed}}
    &  Gemini 2.0 Flash
            & 80.7 \scriptsize{$\pm$ 0.3}  \\
     & GPT-4o 
            & 84.0 \scriptsize{$\pm$ 0.5}  \\
     & Claude 3.5 Sonnet
            & \textbf{84.8 \scriptsize{$\pm$ 0.4}}  \\

    \midrule
    \multirow{2}{*}{\centering\textit{Reasoning}}
    &  GPT-o3-mini 
            & \multicolumn{1}{l}{~~~~66.3}   \\

    &  DeepSeek R1
            & \multicolumn{1}{l}{~~~~77.0} \\
    \bottomrule
  \end{tabular}
  }
  \caption{\label{MCQA_results}
    Average accuracy on \textsc{Morables} over 3 runs ($\pm$ std). Reasoning models are tested only once due to their cost.
  }
\end{table}

Results for the \textsc{Morables} core dataset, shown in Table \ref{MCQA_results}, reveal a pronounced disparity in performance between large and small LLMs.
Larger open models -- Llama 3.3 70B and DeepSeek V3 -- consistently outperform their smaller counterparts, 
with a performance gap of over 40 percentage points between the worst small model, Mistral 7B (28.4\%), and the best large model, Llama 3.3 70B (73.6\%).
Qwen 2.5 7B substantially outperforms other similarly-sized models, though a significant gap remains compared to the larger models.

Among closed models, Gemini 2.0 Flash, Claude 3.5 Sonnet, and GPT-4o achieve the highest accuracies, demonstrating a clear advantage over the best open LLMs. Reasoning-oriented models show mixed results: DeepSeek R1 outperforms its non-reasoning counterpart, while GPT-o3-mini lags behind, likely due to its smaller scale.

\paragraph{Error Analysis}

\begin{table*}
      \setlength{\tabcolsep}{7.75pt}
\scalebox{0.88}{
    \begin{tabular}{lllllll}
    \toprule
    \textbf{Model} & \textbf{GT} & \textbf{Similar-char.} & \textbf{Trait-injected} & \textbf{Feature-based} & \textbf{Partial-story} & \textbf{Invalid} \\
    \midrule
    Mistral 7b        & 28.4 \scriptsize{$\pm$ 0.8} & ~~4.0 \scriptsize{$\pm$ 0.4} & ~~8.0 \scriptsize{$\pm$ 0.3} & 20.4 \scriptsize{$\pm$ 0.6} & 29.3 \scriptsize{$\pm$ 0.7} & 9.8 \scriptsize{$\pm$ 0.9} \\
    Llama 3.1 8b      & 34.0 \scriptsize{$\pm$ 1.5} & 12.9 \scriptsize{$\pm$ 0.8} & 13.2 \scriptsize{$\pm$ 1.2} & 18.1 \scriptsize{$\pm$ 0.5} & 21.7 \scriptsize{$\pm$ 0.5} & 0.0 \\
    Qwen 2.5 7b       & 46.8 \scriptsize{$\pm$ 0.3} & ~~2.6 \scriptsize{$\pm$ 0.2} & ~~6.3 \scriptsize{$\pm$ 0.2} & 16.8 \scriptsize{$\pm$ 0.6} & 27.5 \scriptsize{$\pm$ 1.0} & 0.0 \\
    \midrule
    Mixtral-8x22b     & 56.9 \scriptsize{$\pm$ 1.4} & ~~3.3 \scriptsize{$\pm$ 0.5} & ~~4.0 \scriptsize{$\pm$ 0.4} & 11.5 \scriptsize{$\pm$ 0.4} & 16.2 \scriptsize{$\pm$ 0.6} & 8.0 \scriptsize{$\pm$ 1.1} \\
    Llama 3.3 70B     & 73.6 \scriptsize{$\pm$ 0.6} & ~~2.1 \scriptsize{$\pm$ 0.3} & ~~1.5 \scriptsize{$\pm$ 0.1} & ~~8.8 \scriptsize{$\pm$ 0.4} & 13.1 \scriptsize{$\pm$ 0.9} & 1.0 \scriptsize{$\pm$ 0.1} \\
    DeepSeek V3       & 70.6 \scriptsize{$\pm$ 0.8} & ~~2.5 \scriptsize{$\pm$ 0.2} & ~~2.1 \scriptsize{$\pm$ 0.2} & 10.0 \scriptsize{$\pm$ 0.4} & 14.6 \scriptsize{$\pm$ 0.7} & 0.0 \\
    \midrule
    Gemini 2.0 F.  & 80.7 \scriptsize{$\pm$ 0.3} & ~~1.5 \scriptsize{$\pm$ 0.1} & ~~2.8 \scriptsize{$\pm$ 0.2} & ~~6.8 \scriptsize{$\pm$ 0.1} & ~~8.2 \scriptsize{$\pm$ 0.0} & 0.0 \\
    GPT-4o            & 84.0 \scriptsize{$\pm$ 0.5} & ~~1.6 \scriptsize{$\pm$ 0.1} & ~~1.3 \scriptsize{$\pm$ 0.1} & ~~4.9 \scriptsize{$\pm$ 0.1} & ~~8.1 \scriptsize{$\pm$ 0.5} & 0.0 \\
    Claude 3.5 S.        & 84.8 \scriptsize{$\pm$ 0.4} & ~~2.6 \scriptsize{$\pm$ 0.3} & ~~2.2 \scriptsize{$\pm$ 0.2} & ~~3.7 \scriptsize{$\pm$ 0.2} & ~~6.8 \scriptsize{$\pm$ 0.4} & 0.0 \\
    \midrule
    GPT-o3-mini       & 66.3 & ~~2.7 & ~~3.7 & 10.4 & 16.8 & 0.0 \\
    DeepSeek R1       & 77.0 & ~~2.3 & ~~2.1 & ~~5.6 & 10.4 & 0.0 \\
    \bottomrule
    \end{tabular}
}
\caption{
Average distribution of answer choices over 3 runs (\%, $\pm$ std) for the core \textsc{Morables} dataset. Each column corresponds to a specific type of answer choice: The \textit{GT} (ground truth) reflects accuracy and \textit{Invalid} specifies cases in which the model's output was not in a valid format (after normalization).
}
\label{tab:confusion_matrix}
\end{table*}

Table~\ref{tab:confusion_matrix} presents the choice
{distribution}
for our experiments on the \textsc{Morables} core dataset, showing the average {selection percentage for each choice}.
The ``partial-story'' moral 
is the most frequent incorrect choice across nearly all models,
indicating a primary failure mode where {they} over-rely on initial narrative cues rather than holistically comprehending the entire plot. 
For large models, the ``feature-based'' moral is the second most common error, revealing a notable performance gap between open and closed-source systems. This distractor accounted for 8.8\% of total responses from Llama 3.3 70B, 11.5\% from Mixtral-8x22b, and 10.0\% from DeepSeek V3, considerably higher than those from closed-source counterparts like GPT-4o (4.9\%) and Claude 3.5 (3.7\%).
Smaller models follow the same pattern but exhibit higher error rates across all distractor categories. For example, Llama 3.1 8b is uniquely vulnerable to the ``similar-character'' (12.9\%) and ``trait-injected'' (13.2\%) distractors, suggesting a broader difficulty discriminating among plausible-sounding but incorrect morals. {Overall, this analysis reveals distinct failure modes: larger models are prone to nuanced thematic misinterpretations, while smaller models are 
misled by simpler, superficial cues.}



\subsubsection{TF/NOTO}
Results for the \texttt{TF} and \texttt{NOTO} evaluation variants (detailed in Section \ref{sec:tf_noto}) are presented in Table \ref{TFNOTO_results}, with a granular breakdown of \texttt{TF} metrics in Table \ref{TF_metrics}. 
The accuracy scores for \texttt{TF} (Table \ref{TFNOTO_results}) are generally higher than or comparable to those achieved in the standard evaluation. However, {the \texttt{TF} framing causes} a significant class imbalance (20\% positives, 80\% negatives). Indeed, {the results of Table \ref{TF_metrics} indicate} models tend to over-predict the ``\textit{True}'' class, thereby frequently misclassifying distractors.
{This pattern is reflected in consistently high recall (models identify the correct moral),
but low precision (they mistakenly accept distractors).}

\begin{table}[t!]
  \centering
   \setlength{\tabcolsep}{8pt}
  \scalebox{0.88}{

  \begin{tabular}{lrrr}
    \toprule
    \multirow{2}{*}{\textbf{Model}}                        & \multicolumn{2}{c}{\bf Accuracy \%} & \multirow{2}{*}{\textbf{Cons.}} \\
    \cmidrule(lr){2-3}
               & \multicolumn{1}{c}{{\texttt{TF}}} & \multicolumn{1}{c}{{\texttt{{NOTO}}}} &  \\
    \midrule
    Mistral 7b
        & 31.7 \scriptsize{$\pm$ 0.4}
            & 6.5 \scriptsize{$\pm$ 0.8}
                & 99.3\\
    Llama 3.1 8b
        & 63.4 \scriptsize{$\pm$ 0.1}
            & 14.7 \scriptsize{$\pm$ 1.8}
                & 55.3 \\
    Qwen 2.5 7b
        & 74.1 \scriptsize{$\pm$ 0.3}
            & 26.1 \scriptsize{$\pm$ 1.3}
                & 73.4\\
    \midrule
    Mixtral-8x22b
        & 63.2 \scriptsize{$\pm$ 0.4}
            & 3.1 \scriptsize{$\pm$ 0.1}
                & 85.7\\
    Llama 3.3 70B
        & 77.9 \scriptsize{$\pm$ 0.4}
            & 12.5 \scriptsize{$\pm$ 0.4}
                & 70.1\\
    DeepSeek V3
        & 81.3 \scriptsize{$\pm$ 0.1}
            & 18.4 \scriptsize{$\pm$ 1.5}
                & 50.8\\
    \midrule
    Gemini 2.0 F.
        & 76.0 \scriptsize{$\pm$ 0.1}
            & 29.3 \scriptsize{$\pm$ 0.2}
                & 83.7\\
    GPT-4o 
        & 81.3 \scriptsize{$\pm$ 0.8}
            & \textbf{30.1 \scriptsize{$\pm$ 0.3}}
                & 78.6\\
    Claude 3.5 S.
        & 82.2 \scriptsize{$\pm$ 0.2}
            & 17.3 \scriptsize{$\pm$ 1.3}
                & 62.3\\

    \midrule
    GPT-o3-mini 
        & \multicolumn{1}{l}{\textbf{83.3}}
            & \multicolumn{1}{l}{19.0} 
                & 55.7 \\
    DeepSeek R1
        & \multicolumn{1}{l}{76.1}
            & \multicolumn{1}{l}{25.5}
                & 81.9 \\
    \bottomrule
  \end{tabular}
  }
  \caption{\label{TFNOTO_results}
    Average accuracy ($\pm$ std) over 3 runs for the \textsc{Morables} \texttt{TF} and \texttt{NOTO} variants. The rightmost column (\textit{Consistency}) measures internal consistency, \textit{i.e.}, the percentage of incorrect \texttt{NOTO} choices that models previously labeled as True in the \texttt{TF} task.
  }
\end{table}

\begin{table}[t]
  \centering
  \setlength{\tabcolsep}{6pt}
  \scalebox{0.88}{
  \begin{tabular}{llll}
    \toprule
    \textbf{Model}           
            & \textbf{Prec \%}
                & \textbf{Rec \%}
                    & \textbf{F1 \%}\\
    \midrule
    Mistral 7b
            & 22.3 \scriptsize{$\pm$ 0.1}  
                & \textbf{96.9 \scriptsize{$\pm$ 0.1} }
                    & 36.2 \scriptsize{$\pm$ 0.1}  \\
    Llama 3.1 8b
            & 32.1 \scriptsize{$\pm$ 0.2}  
                & 74.6 \scriptsize{$\pm$ 1.1}  
                    & 44.9 \scriptsize{$\pm$ 0.4}   \\
    Qwen 2.5 7b
            & 41.1 \scriptsize{$\pm$ 0.4}  
                & 68.2 \scriptsize{$\pm$ 0.3}  
                    & 51.3 \scriptsize{$\pm$ 0.4}  \\
    \midrule
    Mixtral-8x22b
            & 34.4 \scriptsize{$\pm$ 0.2}  
                & 92.5 \scriptsize{$\pm$ 0.1}  
                    & 50.1 \scriptsize{$\pm$ 0.3}  \\
    Llama 3.3 70B
            & 47.2 \scriptsize{$\pm$ 0.4}  
                & 89.7 \scriptsize{$\pm$ 0.1}  
                    & 61.8 \scriptsize{$\pm$ 0.4}  \\
    DeepSeek V3 
            & 52.6 \scriptsize{$\pm$ 0.4}  
                & 67.8 \scriptsize{$\pm$ 0.5}  
                    & 59.2 \scriptsize{$\pm$ 0.1}  \\
    \midrule
    Gemini 2.0 Flash
            & 45.1 \scriptsize{$\pm$ 0.1}  
                & 92.5 \scriptsize{$\pm$ 0.1}  
                    & 60.6 \scriptsize{$\pm$ 0.1}  \\
    GPT-4o 
            & 51.8 \scriptsize{$\pm$ 1.3}  
                & 91.6 \scriptsize{$\pm$ 0.8}  
                    & 66.2 \scriptsize{$\pm$ 0.8}  \\
        Claude 3.5 Sonnet
            & 53.4 \scriptsize{$\pm$ 0.1}  
                & 92.1 \scriptsize{$\pm$ 0.4}  
                    &\textbf{67.6 \scriptsize{$\pm$ 0.1}}  \\

    \midrule
    GPT-o3-mini 
            & \textbf{55.8}  
                & 79.1   
                    & 65.5  \\

    DeepSeek R1
            & 45.1  
                & 89.8  
                    & 60.1   \\
    \bottomrule
  \end{tabular}
  }
  \caption{\label{TF_metrics}
    Average classification metrics for the binary \textsc{Morables} \texttt{TF} variant over 3 runs ($\pm$ std).
  }
\end{table}

{
The \texttt{NOTO} variant reveals that models are highly reluctant to select the ``\textit{None of the others}'' option. We hypothesize two reasons for this: \textit{(i)} the distractor options are sensible morals, even if incorrect for the given fable, and \textit{(ii)} the inherent sycophancy of LLMs (\textit{i.e.}, their tendency to agree with user suggestions), an artifact of RLHF finetuning \cite{sharma2024towards}.
To diagnose this behavior, we introduce a \textit{Consistency} metric (rightmost column,
Table \ref{TFNOTO_results}). When a model incorrectly selects a moral instead of \texttt{NOTO}, we compare this choice against its own judgment from the \texttt{TF} setting. A choice is \textit{Consistent} if the model previously labeled that moral as \textit{True}, indicating internal consistency in selecting 
an answer it genuinely believes is valid. Conversely, a choice is \textit{Inconsistent} if previously labeled \textit{False}. This critical finding suggests the model does not believe its own answer is correct, but selects it anyway, revealing a strong aversion to the \texttt{NOTO} option.
Therefore, high consistency suggests errors stem from a stable but flawed thematic interpretation, while low consistency supports the sycophancy hypothesis, as models would rather select a wrong answer than none.}

{The high consistency of Mistral 7b and Mixtral-8x22b reflects their poor performance, evidenced by the large precision-recall gap in Table \ref{TF_metrics}. Llama 3.1 8b's low consistency is initially surprising; however, further analysis shows that it is considerably more likely than Mistral 7b to predict \textit{False} (53\% false predictions compared to only 12\% for Mistral 7b). Notably, Mixtral-8x22b also shows a high rate of False predictions (46.3\%), suggesting Llama 3.1 8b's inconsistency stems from a broader difficulty with moral inference, likely due to its small size.}

Despite their larger sizes, DeepSeek V3 and, to a lesser extent, Llama 3.3 70B exhibit notable inconsistency. DeepSeek V3 often rejects a moral in the \texttt{TF} setting but then selects it around half the time in \texttt{NOTO}, while Llama 3.3 70B does so about 30\% of the time. Remarkably, Claude 3.5 also demonstrates significant inconsistency, changing its stance in roughly 38\% of cases. {Surprisingly, GPT-o3-mini is also highly inconsistent, despite strong \texttt{TF} metrics.}
GPT-4o, DeepSeek R1, and Gemini 2.0 Flash show relatively stable performance across both settings, though a 20\% rate of self-refuted answers remains a non-negligible gap.

\begin{table}[t]
  \centering
   \setlength{\tabcolsep}{10pt}
  \scalebox{0.88}{
  \begin{tabular}{ccccc}
    \toprule

    \textbf{Char.}           
        & \textbf{Adj.} 
            & \textbf{Pre.}
                & \textbf{App.}
                    & \textbf{Acc \%} \\
    \midrule
        \multicolumn{5}{c}{\it {GPT-4o} (baseline accuracy: {84.0} {\footnotesize $\pm$ 0.5})} \\
    \midrule
    
    \checkmark
        &  -
            & -
                &  -
                    &  {82.2} \scriptsize{$\pm$ 0.9} 
                       \\
    
    -    & 1.8\%
            & -
                &  -
                    &  {81.0} \scriptsize{$\pm$ 0.3} 
                         \\
    
    -    &  -
            & 2.8\% 
                &  -
                    &  {79.8} \scriptsize{$\pm$ 0.4} 
                         \\
    
     -   &  -
            & -
                & 4.9\%
                    &  {80.0} \scriptsize{$\pm$ 0.3} 
                        \\
    
   -     &  -
            & 3.2\%
                & 4.6\% 
                    &  {75.7} \scriptsize{$\pm$ 0.1} 
                         \\
    \checkmark
        & 0.9\% 
            & -
                &  -
                    &  {78.6} \scriptsize{$\pm$ 0.2} 
                         \\
    \checkmark
        &  -
            & 4.7\%
                & 5.6\%
                    &  {72.9} \scriptsize{$\pm$ 0.4} 
                         \\
    
    -    &  1.7\%
            & 3.8\% 
                & 3.7\% 
                    &   {74.7} \scriptsize{$\pm$ 0.3} 
                         \\
    \checkmark
        & 1.1\%
            & 5.2\% 
                & 4.9\% 
                    & 71.7 \scriptsize{$\pm$ 0.1} 
                        \\

    \midrule
        \multicolumn{5}{c}{\it Llama 3.3 70B (baseline accuracy: {73.6} {\footnotesize $\pm$ 0.6})} \\
    \midrule

    \checkmark
        &  -
            & -
                &  -
                    &  {72.1} \scriptsize{$\pm$ 0.4} 
                       \\
     
    -    & 1.3\%
            & -
                &  -
                    &  {72.5} \scriptsize{$\pm$ 0.6} 
                         \\
    
    -    &  -
            & 2.2\% 
                &  -
                    &  {73.6} \scriptsize{$\pm$ 0.4} 
                         \\
    
     -   &  -
            & -
                & 9.5\%
                    &  {68.9} \scriptsize{$\pm$ 0.2} 
                        \\
    
   -     &  -
            & 4.2\% 
                & 8.0\% 
                    &  {65.0} \scriptsize{$\pm$ 1.6} 
                         \\
    \checkmark
        & 1.8\%  
            & -
                &  -
                    &  {72.3} \scriptsize{$\pm$ 0.2} 
                         \\
    \checkmark
        &  -
            & 3.6\% 
                & 8.5\% 
                    &  {63.9} \scriptsize{$\pm$ 0.4} 
                         \\
    
    -    &  0.9\% 
            & 4.1\% 
                & 7.4\%
                    &   {65.7} \scriptsize{$\pm$ 0.4} 
                         \\
    \checkmark
        & 0.6\% 
            & 3.3\% 
                & 8.4\% 
                    & 63.1 \scriptsize{$\pm$ 0.8} 
                        \\
    \bottomrule
  \end{tabular}
  }
  \caption{\label{MCQA_mods_results}
    Results for adversarial (\texttt{ADV}) variants of the \textsc{Morables} dataset, showing the average selection rate of the adversarial moral and overall task accuracy ($\pm$ std over 3 runs). Modification types are abbreviated as: \textit{Char.} (character swapping), \textit{Adj.} (adjective injection), and \textit{Pre./App.} (tautology at start/end). Selection rate for \textit{Char.} is omitted, as this modification does not affect distractor morals.
  }
\end{table}

\subsubsection{Adversarial MCQA}
\label{subsec:advmcqa}
Table \ref{MCQA_mods_results} shows the performance of GPT-4o and Llama 3.3 70B on the \textsc{Morables} \texttt{ADV} variants across different modification combinations. 
In general, the modifications lead to a noticeable decline in performance, with their effects worsening when combined. Tautologies prove to be the most impactful modification, particularly when added at the end of the story.


Altering the characters within fables without further changes has minimal impact, with adjective injections yielding similarly modest effects (though slightly more pronounced for GPT-4o).
Inserting tautologies is more impactful, causing a noticeable performance drop as models frequently select them as the answer. The two models also exhibit different positional biases: GPT-4o favors initial tokens, while Llama 3.3 favors final ones.
As expected, combining multiple modifications leads 
to the sharpest decline,
with accuracy dropping by up to 12.3\% for GPT-4o and 10.5\% for Llama 3.3.

\begin{table}[t]
    \centering
      \setlength{\tabcolsep}{13pt}
  \scalebox{0.88}{
    \begin{tabular}{l c c }
        \toprule
        \textbf{Model} & \textbf{Rating} & \textbf{BERTScore} \\ \midrule
        GPT-4o 
            & 3.83 \scriptsize{$\pm$ 1.29} 
                & 0.29  \scriptsize{$\pm$ 0.24}  \\ 
        Claude 3.5 S.
            & 3.76  \scriptsize{$\pm$ 1.29} 
                & 0.24 \scriptsize{$\pm$ 0.21} \\
        Llama 3.3 70B
            & 3.66 \scriptsize{$\pm$ 1.28} 
                & 0.22  \scriptsize{$\pm$ 0.19 } \\
        \bottomrule
    \end{tabular}
    }
    \caption{Average metrics for free-text morals generated by three different LLMs. Annotators rated each moral on a 1–5 scale based on its alignment with the corresponding fable. BERTScore (F1) is also reported.}
    \label{tab:free_text}
\end{table}

\subsection{Free-Text Moral Evaluation}
\label{sec:freetext}
To further contribute to the evaluation of free-text generation, we assess moral generation from three different LLMs (GPT-4o, Claude 3.5, and Llama 3.3 70B). The same annotators involved in our dataset validation separately rated three generated morals per fable on a 1–5 Likert scale, according to the following alignment rubric:
\begin{enumerate}[label=(\arabic*), nosep]
    \item \textit{No alignment}: The moral does not relate to the story's core message;
    \item \textit{Minimal alignment}: The moral is related to the story but misses the main point;
    \item \textit{Partial alignment}: The moral captures a part of the story's message;
    \item \textit{Strong alignment}: The moral is a good fit but doesn't capture the full essence;
    \item \textit{Perfect alignment}: The moral perfectly encapsulates the story's essence.
\end{enumerate}

Table \ref{tab:free_text} summarizes our results, including the BERTScore \cite{BERTScore} F1 similarity between each generated moral and the reference for an approximate measure of semantic similarity.

All three models produce morals that, according to annotators, generally align with their respective fables. However, the high standard deviation 
suggests considerable variability, likely reflecting the inherent subjectivity of interpreting morals and the challenges of evaluating 
open-ended free-text outputs. Meanwhile, BERTScore F1 values remain fairly low for all models, indicating that the wording of generated morals often diverges substantially from the original references. 
A qualitative analysis of the divergence between BERTScore and human ratings is provided in Appendix \ref{app:user_study}.




To explore the link between automated similarity measures and human judgment, we calculated the Pearson correlation coefficient between BERTScore and average annotator ratings. The correlation coefficients are positive albeit relatively low -- 0.35 for Claude 3.5, 0.28 for GPT-4o, and 0.33 for Llama 3.3 -- indicating only a weak positive relationship between semantic similarity and human perception of alignment (see the plot in Appendix \ref{app:user_study}).
This suggests that automated semantic similarity scores are unable to fully capture the complexity of the task, a finding that aligns with previous work on the limited agreement between such metrics and human judgment \cite{leung-etal-2022-semantic,herbold2024semantic}

\subsection{Discussion}

Our findings reveal several noteworthy patterns. First, smaller models struggle with the core moral inference task, emphasizing its non-trivial nature. In contrast, large open models such as Llama 3.3 70B perform well, although closed-source models still maintain a significant advantage.

Analysis of the \texttt{TF} and \texttt{NOTO} variants reveals that models frequently regard multiple options as plausible, often selecting the least incorrect option, echoing observations by \citet{wang-etal-2025-llms-may}. This is reflected in the models' high recall but low precision in the \texttt{TF} setting (Table \ref{TF_metrics}) and diminished accuracy in the \texttt{NOTO} scenario.
Notably, models exhibit inconsistent behavior, sometimes preferring to choose a moral they previously rejected in \texttt{TF} over a ``\textit{None of the others}'' option.  
This may reflect the propensity of LLMs to prioritize internal probability rankings over strict correctness, making ``\textit{None of the others}'' rarely the preferred selection when juxtaposed with plausible alternatives \cite{wang-etal-2025-llms-may}. As noted by \citet{salido2025othersgeneraltechniquedistinguish}, this may also signal strong memory-based associations.

Reasoning models yield mixed results: DeepSeek R1 almost matches closed-source systems, while GPT-o3-mini performs notably worse -- both in terms of accuracy on the core dataset as well as in consistency between \texttt{TF} and \texttt{NOTO} -- illustrating pronounced limitations. This suggests that model scale remains more important than reasoning capabilities for this task.

The impact of adversarial modifications may also indicate memorization, as evidenced by the significant performance drop in GPT-4o and Llama 3.3. Both models show a tendency to focus on the initial or final tokens of narratives, which heavily influences their decisions. This aligns with previous findings that neural language models often exhibit ``attention sinks'' at the beginnings and ends of documents \cite{xiao2024efficient,zhang2024found}.

\section{Conclusion \& Future Work}
We introduce \textsc{Morables}, a high-quality, manually curated benchmark of fables paired with their morals.
By building several challenging tasks, we probe the nuanced reasoning and narrative understanding capabilities of LLMs.
Our results reveal a substantial
performance gap between small and large models, with only the largest models achieving strong accuracy. However,
even large models display inconsistent behavior under specialized evaluation, such as the \texttt{TF} and \texttt{NOTO} variants, and remain vulnerable to adversarial modifications that expose reliance on memorization and positional biases rather than genuine moral inference. These findings highlight important challenges for future research.
Future work will expand the dataset to include fables from diverse cultures, facilitating the study of cultural bias and differing ethical perspectives. 
Moreover, we plan to leverage the holistic nature of moral lessons -- which often arise from the broader narrative rather than specific textual elements --
as a testbed for explainability research. 

\section*{Limitations}
The primary limitation of \textsc{Morables} lies in the high likelihood that most modern LLM have encountered not only the stories but also their associated morals during their pre-training phase.  This raises significant concerns about memorization, which could compromise our ability to draw definitive conclusions regarding their moral inference capabilities, as the models may simply be recalling information they have already seen. While our adversarial modifications aim to partially address this issue, a more effective (but resource-intensive) solution would be to create entirely new, unseen short stories to test whether LLMs can still infer a moral in those cases.

Another limitation is that the dataset is predominantly sourced from Western stories. This is largely due to the greater availability of English fables that come with verified morals. Consequently, this focus restricts our ability to study potential cultural biases. Likewise, the dataset and our evaluation was performed for English only.
Lastly, the rapid release of new models makes it challenging to publish results for models developed concurrently with ongoing experimentation. However, while future work should benchmark newer models, significant effort will be required to determine whether any observed performance improvements stem from enhanced reasoning capabilities or increased memorization tendencies.

Finally, we acknowledge a methodological consideration regarding GPT-4o's evaluation. Since the model's outputs were used to generate the distractor choices for our dataset, there is a risk of ``self-preferential bias'' in its results on the MCQA benchmark. However, we found no clear evidence of such an effect: GPT-4o's performance was not an outlier compared to its peers in a way that would indicate an unfair advantage. Furthermore, the risk was inherently limited because the model was not exposed to the original fable's text during the choice-generation process. Therefore, while the potential for bias is noted, its inclusion remains valuable for comparative analysis.


\section*{Ethics Statements and Broader Impact}
\textsc{Morables} was collected from public web resources under licenses that allow use and redistribution for research purposes. The text and the moral of the fables has been extracted and formatted by the authors of this work. Human annotators were hired to verify the quality of the dataset and of AI-generated morals. Annotators were asked to read the fables and determine the correct moral from five multiple choices, as well as to rate automatically generated morals on a Likert scale from 1 to 5. Though all fables go through an initial screening process for offensive content, annotators were allowed to skip a fable if it contained content they felt uncomfortable with and encouraged to leave an explanatory comment. 
Annotators were paid around £20 per hour, well above the minimum hourly wage in the UK, where annotation took place.
Annotators are recruited from diverse backgrounds, including Computer Science, Neuroscience, Theology, Literature.

As discussed in the previous section, our research focuses exclusively on fables from the Western tradition that have been translated into English. This choice is largely influenced by the prevalence of fable-moral pairs in English literature. However, fables are a universal phenomenon found in cultures around the world, and nearly all human societies utilize them. As a result, our dataset is unfortunately more than likely biased toward moral lessons that reflect Western values. This limitation may pose two significant issues: first, it fails to adequately represent values that differ across cultures, and second, models trained on this dataset may inadvertently reinforce these biases.
Future work should aim to broaden this scope to include fables from diverse cultures globally. It would be particularly interesting to explore how models respond to fables with messages that diverge from conventional Western themes.

\section*{Acknowledgments}

We thank our team of annotators --  Felix Chadwick-Smith, Fayeja Farhana, Dhruvil Raval, Kayleigh Shier, and Hannah Trotman -- for their careful and thoughtful work in validating our dataset and rating the model-generated morals. Our sincerest gratitude also goes to the Cardiff NLP group for their support throughout the project and for their detailed review of an early manuscript.

Jose Camacho-Collados was supported by a UKRI Future Leaders Fellowship.

\bibliography{anthology, custom}

\clearpage
\appendix
\section{Dataset Sources and Feature Analysis}
\label{sec:appendix_dataset}



\subsection{Data sources}
Table \ref{tab:data_sources} lists websites crawled for the dataset collection. 
All stories are free to use and redistribute for research purposes.
Aesop's fables originated from an oral tradition, making it difficult to determine whether some fables attributed to other authors are direct transcriptions or new works inspired by him.
Notable sources of such fables in this study include the Roman fabulist Phaedrus and the Italian writer Laurentius Abstemius. Whenever possible, we utilize English translation by G. F. Townsend \cite{aesop2008} and L. Gibbs \cite{gibbs2017}\footnote{The author kindly granted permission to use the content published under a Creative Commons license.}.

All morals used in our benchmark are directly sourced from either the original authors or reputable later authors/translators who provided interpretations of these fables:
\begin{itemize}[nosep]
    \item \textit{Original Authors/Translators}: Some morals were explicitly added by the original authors of the fables (\textit{e.g.}, Aesop) and subsequently translated.
    \item \textit{Later Authors/Fabulists}: In many cases, morals were appended by later authors or fabulists who re-interpreted and translated the fables. For example, for a fable like ``The Wolf and the Raven'' (Perry Index 190), Aesop did not originally provide a moral. However, Sir Roger L'Estrange later added an ``epimythium'' (a moral appended at the end of the story) in his 1692 translation.
\end{itemize}

\subsection{Analysis of Dataset Features}


In addition to standard metrics reported in Section \ref{sec:benchmark}, we measure the Flesch-Kincaid readability score \cite{kincaid1975derivation}. The Fables average a score of 8.30 (with a median of 7.90), indicating that the text is suitable for someone at the 8th-grade reading level. This means that a typical 13- to 14-year-old should be able to read and understand the text with relative ease. A subset of our dataset, attributable to fables by Phaedrus, have a slightly more complex language, with a mean score of 11.59.

The most common characters in the fables are animals, with the fox (70 occurrences), lion (62), dog (61), wolf (52), donkey (36), and eagle (29) being the most frequent. Men are prominently portrayed, often interacting with animals, appearing at least 183 times as ``man'' and many other times in forms such as farmer (26) and shepherd (22). Major deities from Greek and Roman traditions, such as Zeus and Jupiter, also appear prominently, each featured in 20 fables.

We perform an exploratory topic analysis of the dataset with BERTopic \cite{grootendorst2022bertopic}, specifically focusing on the content of the morals. 
An unguided clustering reveals numerous categories which can be loosely categorized as discussions of values and ethics (trust and deception, justice, gratitude, kindness), mindfulness and awareness (life choices and their impact), appreciation (contentment and value, true worth), protection (caution against enemies, suffering and neglect), and facing the human experience with resilience (fate and consequences, courage in adversity).


\begin{table*}[t]
    \centering
    \begin{tabular}{l l l}
    \toprule
    Source name & Author / Translator & Source link \\
    \hline
         \textit{The Aesop for Children }
            & Aesop
                & \href{https://www.gutenberg.org/files/19994/19994-h/19994-h.htm}{Project Gutenberg} \\
         \textit{Aesop Fables }
             & Aesop / G. F. Townsend  
                &  \href{https://www.gutenberg.org/files/21/21-h/21-h.htm}{Project Gutenberg} \\
         \textit{The fables of Aesop}
            & Aesop /  Croxall, La Fontaine, L'Estrange
                & \href{https://babel.hathitrust.org/cgi/pt?id=hvd.hwm65b}{Hathi trust}\\
         \textit{The Comedies of Terence [...]}
            & Terence, Phaedrus / H. T. Riley
                &\href{https://www.gutenberg.org/files/25512/25512-h/25512-h.htm}{Project Gutenberg} \\
        \textit{Aesop's Books}
            & Aesop / L. Gibbs 
                & \href{https://aesopsbooks.blogspot.com/}{Author Blog}* \\
         \bottomrule
    \end{tabular}
    \caption{List of source web pages used for constructing \textsc{Morables}. * Permission to utilize the translations and a parsable XML of the website has been kindly granted by the author.}
    \label{tab:data_sources}
\end{table*}

\subsection{Tautologies}
Table \ref{tab:tautologies} presents examples of tautologies utilized for the \texttt{ADV} variant of \textsc{Morables}. The full set of tautologies is provided in the linked repository.

\begin{table*}[t]
    \centering
    \begin{tabular}{ll}
    \toprule
    \textbf{Tautology} & \textbf{Moral} \\
    \hline
    \textit{It is what it is.} 
        & Accept things as they are. \\
    \textit{What is, is. }
        & Embrace the present reality. \\
    \textit{The outcome is the outcome.} 
        & Accept what results from your endeavors. \\
    \textit{Things are what they are. }
        & Accept the finality of reality without resistance. \\
    \bottomrule
    \end{tabular}
    \caption{Examples of tautologies and paired morals utilized for the \texttt{ADV} version of \textsc{Morables}.}
    \label{tab:tautologies}
\end{table*}

\section{Opposite morals and binary task}
\label{sec:appendixBinary}
We considered other distractors for the core dataset, such as semantically opposite generated morals with high word overlap and random sentences from the fable to check for selection biases. However, these options were almost never chosen by the models in preliminary tests, indicating they were too easy and leading to their exclusion. Moreover, including a semantically opposite negative option might have inadvertently provided a superficial cue to the correct answer by allowing models (or even human evaluators) to simply identify the contrasting pair, thereby reducing the choice to just those two options.
For completeness, we also implement a binary task where models select between the original and the opposite moral, which we detail below.

\paragraph{Preliminary binary task}
Following the work of \citet{guan-etal-2022-corpus}, we devise a preliminary task consisting of a binary classification between two moral choices. The task is to differentiate between the original and its semantic opposite.
While \citet{guan-etal-2022-corpus} create the incorrect candidate by substituting a random token in the original moral with its antonym, we follow a different approach. Leveraging the strength of LLMs, we prompt GPT-4o to provide ``anti-morals'' for each provided moral, stressing the fact that the provided answer should be semantically opposite in a binary sense and share as many words as possible as the original moral. The morals are collected in a 3-shot scenario, where examples are hand-crafted. 
The resulting opposite morals are semantically and linguistically consistent and coherent. Nevertheless, the expectation remains for this binary task to be easy, as morals usually encapsulate sensible and just ethical values. Moreover, morals in themselves are often popular proverbs. 
\begin{table}[t]
  \centering
  \scalebox{0.88}{\begin{tabular}{lccc}
    \toprule
    \textbf{Model}           & \textbf{Acc}\textbf{ \% }& \textbf{Acc$_\emptyset$} \textbf{\% } & \textbf{Acc$_{1S}$} \textbf{\%}\\
    \hline
    Llama 3.3 70B
        & 94.8
            & 90.0
                & 87.7 \\
    Gemini 2.0 Flash
        & 96.9 
            & 94.2 
                & 92.8  \\
    GPT-4o 
        & 97.6
            & 89.8 
                & 95.3  \\
    GPT-o3-mini 
        & 95.5 
            & - 
                & -  \\
    DeepSeek V3
        & 97.6
            & 91.7
                & 90.8 \\
    DeepSeek R1
        & 92.7 
            & 91.3 
                & 89.8 \\
    \bottomrule
  \end{tabular}}
  \caption{\label{Binary_results}
    Accuracy for the \textsc{Morables} Binary preliminary task. \textit{Acc} stands for accuracy in the standard setting, while \textit{Acc$_\emptyset$} and \textit{Acc$_{1S}$} stand for accuracy when the model is provided no fable or just the first sentence of the fable, respectively. 
  }
\end{table}

\paragraph{Results of preliminary tests}
Results for the binary task are presented in Table \ref{Binary_results}. As expected, opposite morals can be easily distinguished, notably even without providing the fables to the models. This is demonstrated by \textit{Acc$_\emptyset$}, which measures model accuracy when determining the correct choice without any story text. We also report \textit{Acc$_{1S}$}, where only the first sentence of the fable is provided. Interestingly, performance slightly decreases in this case, suggesting that models place considerable emphasis on the task's initial tokens and not just the choice answers.


\section{User study}
\label{app:user_study}
\subsection{Discrepancies between BERTScore and Human Ratings}

Figure \ref{fig:pearson} shows the weak positive link between semantic similarity (BERTScore) and average human ratings, indicating that greater semantic overlap does not strongly predict human preference in moral evaluation. 

A low BERTScore with a high human rating often occurs when the original moral is abstract and the model presents instead a more explicit lesson.
Occasionally, however, this may result in the generated moral becoming overly focused on story details. For instance, the moral ``\textit{Straws show how the wind blows}'' (Perry Index 95) is an abstract lesson on how small details can reveal important clues about future events or the character of an individual. In this case, the generated moral ``\textit{The way you are regarded by strangers will reflect how those closer to you perceive you}'' fits the narrative of the story, but fails to encapsulate its broader meaning e.

Conversely, a high BERTScore does not always guarantee a high human rating.
For instance, Figure \ref{fig:examplediff}
illustrates a case where all model-generated morals are semantically similar and aligned with the original moral, thus being correct. However, users favored more detailed, discursive versions like Claude's. 
We hypothesize this discrepancy signals an annotation artifact, where morals are evaluated comparatively rather than independently, despite instructions to assess each separately.

\begin{figure}[ht] 
    \footnotesize
    \centering
    \scalebox{0.95}{\begin{contentbox}[title={The Two Roosters And The Eagle (Perry 281)}]
            
            Two roosters were fighting with one another. The loser hid himself away in a corner, while the rooster who had won the battle flew up on top of the house and flapped his wings, crowing about his victory. An eagle then swooped down and carried the rooster away.

        \tcblower
         
        \begin{itemize}[itemsep=2pt, parsep=0pt, left=0pt] 
            \item \textbf{True Moral}: Pride goes before destruction.
            \item \textbf{GPT}: Pride comes before a fall.
            \item \textbf{Claude}: The proud are often brought low.
            \item \textbf{Llama}: Pride goes before a fall.

        \end{itemize}
    
    \end{contentbox}
    }

    \caption{Example of a Fable with its original moral and three LLM-generated morals. Despite all morals being semantically similar and correct, users often preferred more discursive and elaborated morals, such as the one generated by Claude.}
    \label{fig:examplediff}
\end{figure}

\begin{figure*}
    \centering
    \includegraphics[width=0.85\linewidth]{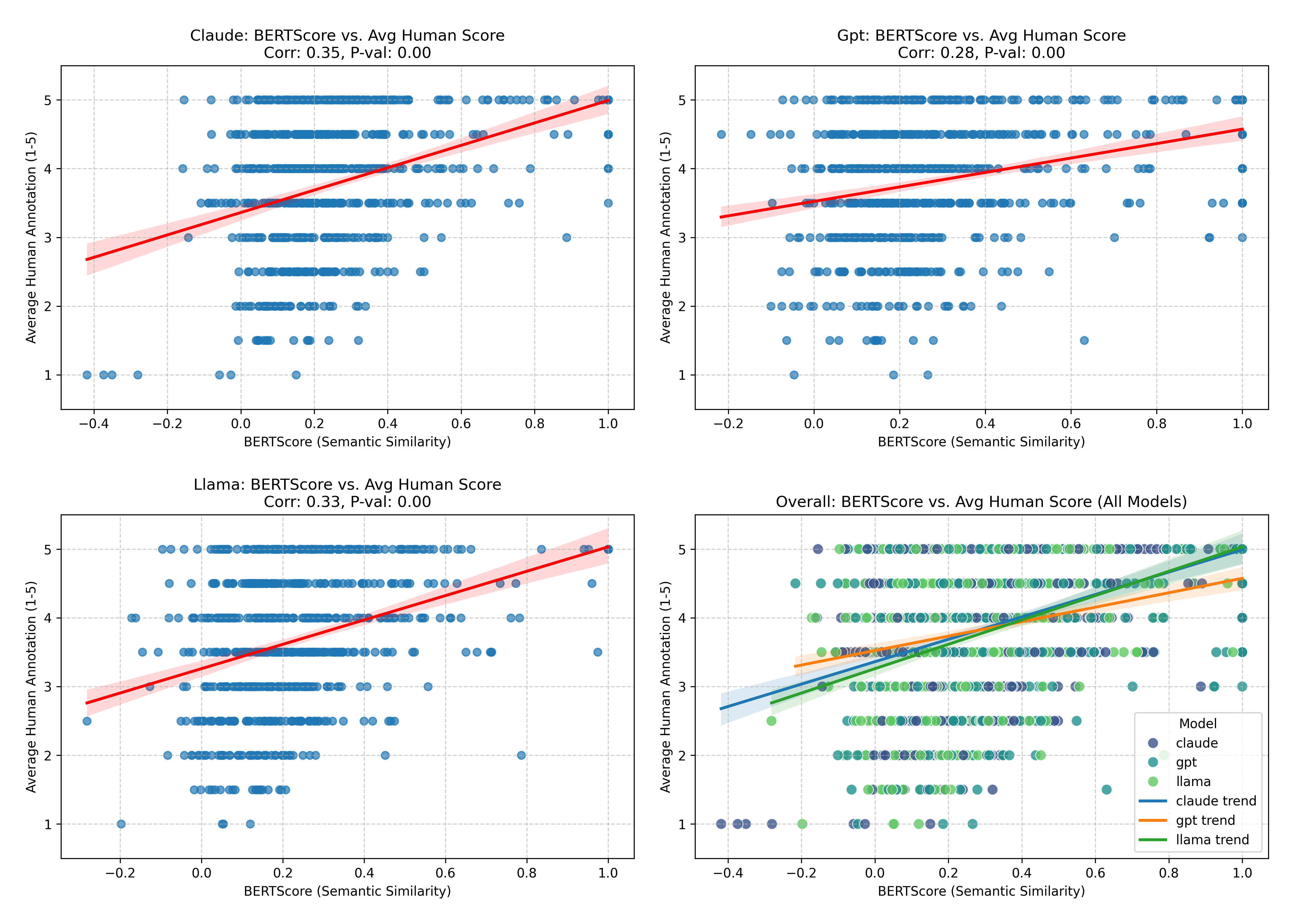}
    \caption{Average Human Rating vs. BERTScore across three LLMs. All models exhibit only a weak positive correlation, highlighting both the difficulty of evaluating morals using semantic similarity metrics and the inherent subjectivity in human moral assessment.}
    \label{fig:pearson}
\end{figure*}




\subsection{Annotation Graphical User Interface}
Figure~\ref{fig:gui_multifig} showcases the two components of the annotation user interface used in this study: \textit{(a)} the MCQA component, in which annotators select the appropriate moral(s) for a given fable, and \textit{(b)} the moral evaluation component, where annotators rate the alignment of each moral with the fable.

\begin{figure*}[ht]
    \centering
   \scalebox{0.95}{ \begin{subfigure}[b]{0.95\textwidth}
        \centering
        \fbox{\includegraphics[width=\textwidth]{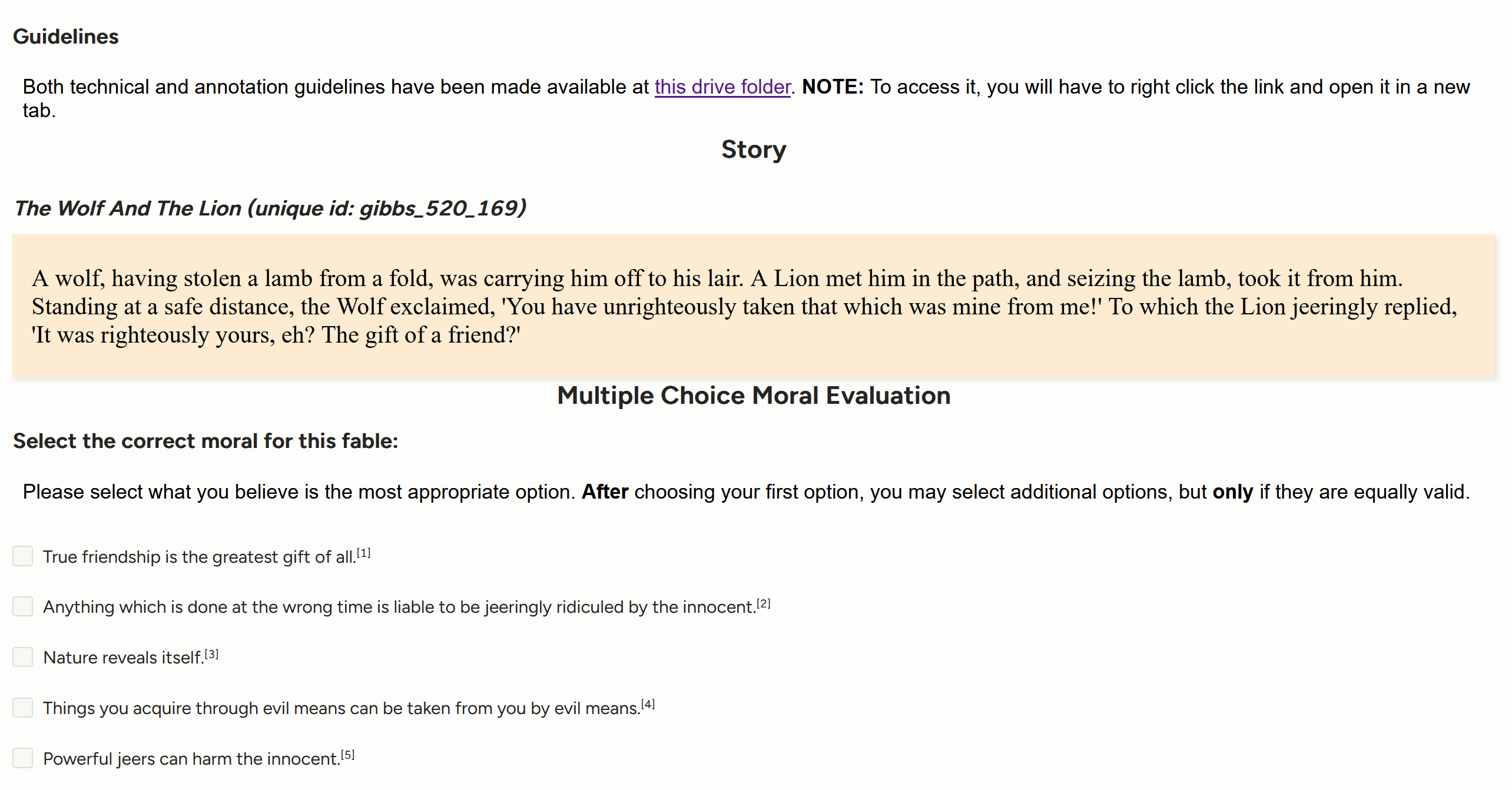}}
        \caption{MCQA part of the graphical user interface.}
        \label{fig:gui1}
    \end{subfigure}}
    \vspace{0.5em}
    \scalebox{0.95}{\begin{subfigure}[b]{0.95\textwidth}
        \centering
        \fbox{\includegraphics[width=\textwidth]{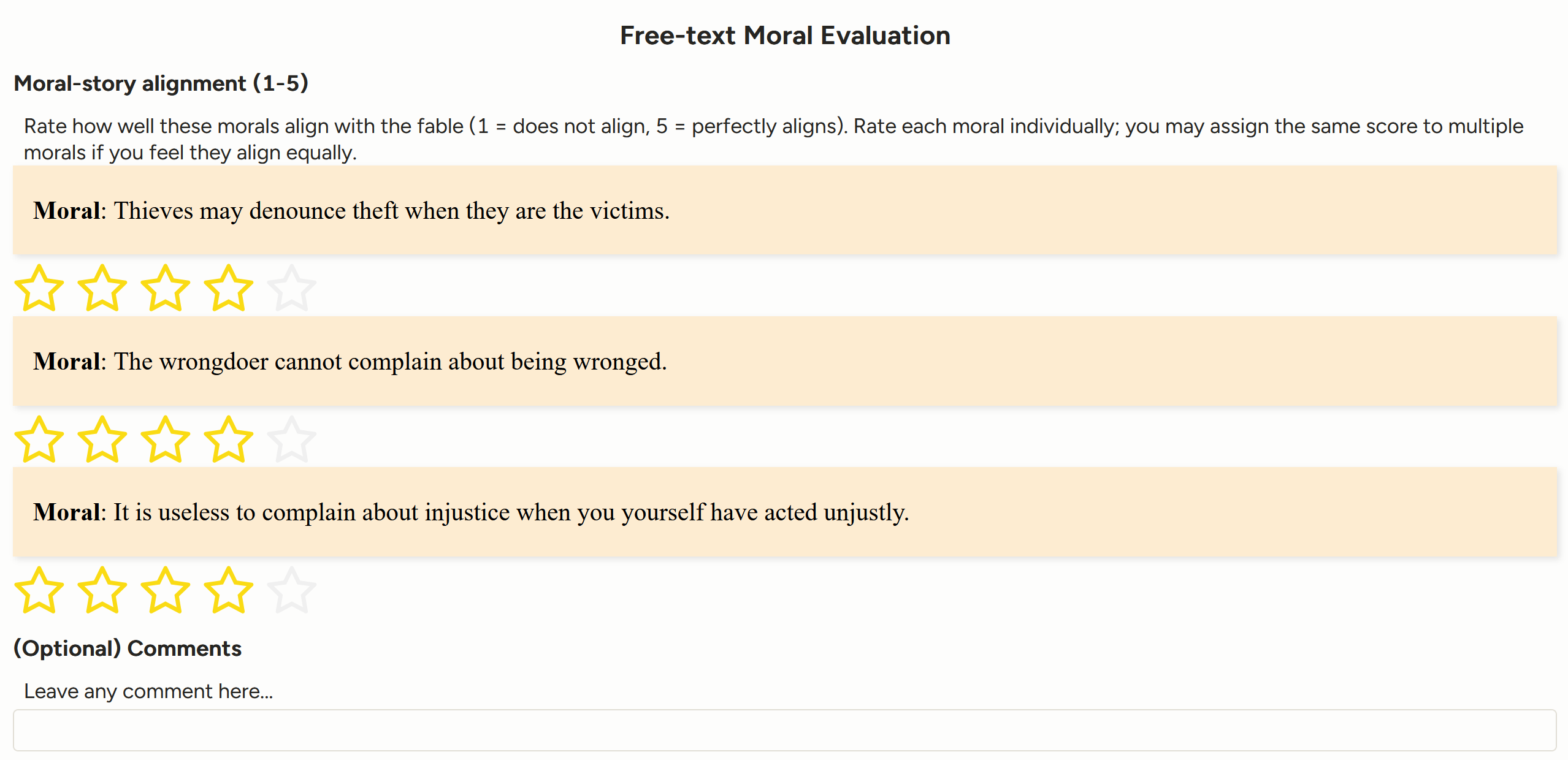}}
        \caption{Free-text moral evaluation part of the graphical user interface.}
        \label{fig:gui2}
    \end{subfigure}}
    \caption{Annotation user interface used in this study. \textit{(a)} MCQA component, where annotators read a fable and select the appropriate moral(s); multiple selections are permitted to capture ambiguity. \textit{(b)} Moral evaluation component, where annotators rate each of three provided morals according to their alignment with the fable, using a Likert scale from 1 (does not align) to 5 (perfectly aligns).}
    \label{fig:gui_multifig}
\end{figure*}
\section{Empirical results on output tokens}
\label{sec:appendix_output_tokens}

Below, we present empirical results on several relevant phenomena identified in the literature that may skew LLM MCQA evaluation outcomes. While not exhaustive, these experiments provide insights and highlight areas for future investigation.

\subsection{Output selection}
We examine how model performance changes when using the first generated token as the answer (Acc$_{NT}$) compared to using output probabilities for option ID tokens (Acc$_{LP}$), as shown in Table \ref{tab:nextvslps}. The latter method is applicable to open, local models and can be partially implemented via OpenAI's API for GPT-4o, which provides log probabilities for the 20 most likely tokens; in practice, answer labels are consistently among these top 20. However, this functionality is not supported by other models such as Gemini Flash 2.0, Claude 3.5, and reasoning models (GPT-o3-mini and DeepSeek R1).

\begin{table}[t]
  \centering
  \setlength{\tabcolsep}{9pt}
 \scalebox{0.88}{ \begin{tabular}{lcc}
    \toprule
    \textbf{Model}           
        & \textbf{Acc$_{NT}$ \%}  
            & \textbf{Acc$_{LP}$ \%} \\
    \hline
    Llama 3.3 70B
        & 70.0 \scriptsize{$\pm$ 0.6}  
            & 71.0 \scriptsize{$\pm$ 1.1}  \\
    GPT-4o 
        & 83.0 \scriptsize{$\pm$ 1.7}  
            & 81.5 \scriptsize{$\pm$ 2.1}  \\
    \bottomrule
  \end{tabular}}
  \caption{\label{tab:nextvslps}
    Accuracy on a reduced sample (100 data points) of the \textsc{Morables} dataset. Acc$_{NT}$ stands for accuracy utilizing the standard next token prediction evaluation, while Acc$_{LP}$ indicates the accuracy for the same run evaluated using log probabilities ($\pm$ std).
  }
\end{table}

\begin{table*}[t]
  \centering
  \setlength{\tabcolsep}{12pt}
  \scalebox{0.88}{\begin{tabular}{lccccc}
    \toprule
    \textbf{Model}           
        & {O$_1$}
        & {O$_2$}  
        & {O$_3$}  
        & {O$_4$}  
        & {O$_5$} \\
    \hline
    Core dataset GT distribution
        & 20.17\%
        & 21.16\%
        & 19.46\%
        & 19.04\%
        & 20.17\% \\
    \hline
    Llama 3.3 70B
        & \textbf{16.92\% }
        & \textbf{25.53\%}
        & 22.43\%
        & 17.21\%
        & 17.91\% \\
    GPT-4o 
        & 20.59\%
        & 19.46\%
        & 20.59\% 
        & 19.18\%  
        & 20.17\% \\
    Gemini 2.0 Flash
        & 20.29\%
        & 18.84\%
        & 22.90\%  
        & 19.13\%  
        & 18.84\% \\
    DeepSeek V3
        & 22.28\%
        & 22.14\% 
        & 21.02\% 
        & 20.31\%  
        & \textbf{14.25\%} \\
    \bottomrule
  \end{tabular}}
  \caption{\label{tab:tokenbias}
        Answer selection percentages on a sample (50\%) of the \textsc{Morables} dataset. Ground Truth distributions are shown in the first row. Percentages that vary significantly are highlighted in bold.
  }
\end{table*}

We conducted three runs on a reduced sample of 100 data points to assess whether the choice of evaluation procedure significantly affects the results. This analysis is relevant only when the model temperature is set above 0, as otherwise models yield identical outputs regardless of the evaluation method. Accordingly, we set $T=1$ for our evaluation. For each run, we calculate accuracy by comparing the next generated token as the answer to the top answer based on log probabilities within the same run.

We evaluate one open model (Llama 3.3) and one closed model (GPT-4o). As shown in Table \ref{tab:nextvslps}, the confidence intervals for the two approaches overlap, indicating that both are valid alternatives without a statistically significant advantage for either model.

\subsection{Token bias}
We investigate the presence of token bias in LLMs for our benchmark in a selection of models \cite{zheng2024large,wei-etal-2024-unveiling}. 
Results are summarized in Table \ref{tab:tokenbias}. Overall, the token bias phenomenon manifests with varying magnitudes across different models, with advanced models like GPT-4o showing little to no evidence of token bias, while models such as Llama 3.3 and DeepSeek V3 (and, to a lesser extent, Gemini 2.0 Flash) exhibit potential token bias in their selection of certain choices over others. Specifically, Llama 3.3 appears to favor the second option, whereas DeepSeek V3 tends to select the last option less frequently. This behavior occurs regardless of the answer ID naming conventions, which are discussed in the next section.

\subsection{Token naming}
We also evaluate whether models have different performance based on whether the answer choices are named with letters (``A'', ``B'', \textit{etc.}) or numbers (``1'', ``2'', \textit{etc.}). Results are summarized in  Table \ref{tab:tokennaming}. Similarly to the token bias phenomenon, the results depend on the model, with GPT-4o showcasing basically no difference, while models like Llama 3.3 and even Gemini 2.0 Flash display a small difference in performance. Interestingly, Gemini performs better on letter-named choices, while Llama performs worse. In general, there appears to be no strictly better approach to choice naming in our case. 

\subsection{Number of choices}
Table \ref{DiffChoic} compares model performance when presented with more than five answer choices (specifically, eight, due to the inclusion of adversarial options) versus when limited to five choices, in which less effective distractors were removed. The reduction to five choices was achieved by eliminating distractors that the model had not previously selected as correct answers. The results indicate that performance remains largely consistent across both settings, with a slight improvement in the five-choice condition, particularly for GPT-4o.  
This suggests that while the number of choices may slightly impact performance, the nature and quality of the distractors play a more important role.

\begin{table}[t]
  \centering
  \setlength{\tabcolsep}{9pt}
  \scalebox{0.88}{\begin{tabular}{lcc}
    \toprule
    \textbf{Model}           
            & \textbf{Acc$_8$ \%}
                & \textbf{Acc$_5$ \%}\\
    \midrule

    Llama 3.3 70B
            & 63.1 \scriptsize{$\pm$ 0.8} 
              & 63.9 \scriptsize{$\pm$ 0.5}  \\
    GPT-4o 
            & 71.7 \scriptsize{$\pm$ 0.1} 
               & 73.1 \scriptsize{$\pm$ 0.7}  \\
    \bottomrule
  \end{tabular}}
  \caption{\label{DiffChoic}
  Accuracy on the most challenging \texttt{ADV} variant of the \textsc{Morables} dataset, which incorporates all modifications. The table compares performance on more than five choices (left) versus five (right), where the latter set excludes the least plausible distractors.
  }
\end{table}

\begin{table}[t]
  \centering
  \setlength{\tabcolsep}{7pt}
  \scalebox{0.88}{\begin{tabular}{lcc}
    \toprule
    \textbf{Model}           
        & \textbf{Acc$_{123}$ \%}  
            & \textbf{Acc$_{ABC}$ \%} \\
    \hline
    Llama 3.3 70B
        & 73.0 \scriptsize{$\pm$ 0.6}  
            & 70.0 \scriptsize{$\pm$ 1.1}  \\
    GPT-4o 
        & 82.5 \scriptsize{$\pm$ 2.1}  
            & 81.5 \scriptsize{$\pm$ 0.7}  \\
    Gemini 2.0 Flash
        & 74.0  \scriptsize{$\pm$ 1.1}  
            & 77.0 \scriptsize{$\pm$ 0.6}  \\
    \bottomrule
  \end{tabular}}
  \caption{\label{tab:tokennaming}
     Accuracy comparison on a sample (100 data points) of the \textsc{Morables} core dataset, showing performance when token IDs are represented as numbers versus as letters ($\pm$ std).
  }
\end{table}

\section{Experimental Setup and Prompts}
\label{sec:appendix_prompts}

\subsection{Implementation Details}
To ensure consistent and reproducible access, all models were queried via the OpenRouter API \cite{openrouter}.

\subsection{Zero-shot vs Few-shot}
Table \ref{shot_comparison} displays the comparison between 0- and 1-shot setting performance for the \textsc{Morables} dataset. Smaller models are excluded from this analysis, as they do not adhere to the format instructions without providing examples. 

The performance gap is notable across most models, with a decrease in accuracy of approximately 4 to 6\% when no examples are given (the substantial drop in performance for DeepSeek V3 is primarily due to a significant number of invalid response formats). Interestingly, reasoning models seem to be less susceptible to the lack of examples.

\subsection{Prompts utilized}
For reproducibility and clarity, all prompts employed in this study are provided at the end of this document, each clearly labeled with the task to which it pertains.

\newpage

\begin{table}[hbt]
  \centering
  \setlength{\tabcolsep}{6pt}
  \scalebox{0.88}{
  \begin{tabular}{lccc}
    \toprule

    \multirow{2}{*}{\textbf{Model}} 
        & \multicolumn{3}{c}{\bf Accuracy \%} \\
        \cmidrule(lr){2-4}
               & \textbf{0-shot \%}
                & \textbf{1-shot \%} & $\Delta_{0-1}$   \\
    
    \midrule

    Llama 3.3 70B
            & 70.4 \scriptsize{$\pm$ 0.2}  
              &  73.6 \scriptsize{$\pm$ 0.6}
                & 3.2 \scriptsize{± 0.8}  \\

    Gemini 2.0 Flash
            & 76.1 \scriptsize{$\pm$ 0.2}  
                & 80.7 \scriptsize{$\pm$ 0.3} 
                & 4.6 \scriptsize{± 0.5} \\
    DeepSeek V3 
            & 50.4 \scriptsize{$\pm$ 1.8}  
                & 70.6 \scriptsize{$\pm$ 0.8} 
                & 20.2 \scriptsize{± 2.6} \\
    Claude 3.5 Sonnet
            & 81.0 \scriptsize{$\pm$ 0.3}  
                & 84.8 \scriptsize{$\pm$ 0.4} 
                & 3.8 \scriptsize{± 0.7} \\
    GPT-4o 
            & 78.0 \scriptsize{$\pm$ 0.4}  
                & 84.0 \scriptsize{$\pm$ 0.5} 
                & 6.0 \scriptsize{± 0.9} \\

    \midrule
    GPT-o3-mini 
            & 65.7
                & 66.3  
                & 0.6 \\

    DeepSeek R1
            & 74.3 
                & 77.0 
                & 2.7 \\
    \bottomrule
  \end{tabular}
  }
  \caption{\label{shot_comparison}
    Accuracy comparison between 0-shot and 1-shot performance for  \textsc{Morables} over 3 runs ($\pm$ std). $\Delta_{0-1}$ indicates the net change in performance when changing between the two settings.
  }
\end{table}

\begin{figure*}
    \footnotesize

    \begin{prompt}[title={Opposite Moral Generation}]
    You are a helpful AI specializing in analyzing the morals of fables and short stories. When given a moral, provide an anti-moral that conveys the opposite meaning. The anti-moral should be semantically opposite in a binary sense and share as many words as possible with the original moral.
    Ensure that the anti-moral maintains a similar length and language style to the original moral, using the same words whenever feasible.
    
    /* Examples */ 
    
    Example 1:
    
    Moral: ``Honesty is the best policy.''
    
    Anti-moral: ``Dishonesty is the best policy.''
    
    Example 2:
    
    Moral: ``An ounce of prevention is worth a pound of cure.''
    
    Anti-moral: ``An ounce of cure is worth a pound of prevention.''
    
    Example 3:
    
    Moral: ``Wealth is of little value if one is too afraid to use or enjoy it.''
    
    Anti-moral: ``Wealth is of great value if one is too afraid to use or enjoy it.''
    \end{prompt}
\end{figure*}

\begin{figure*}
    \footnotesize
    \begin{prompt}[title={Character Alternatives Extraction}]
     You are a helpful AI specializing in extracting information from fables and short stories. Analyze the provided text to identify all characters (main and minor) and important objects. Then, suggest two alternatives for each, ensuring that:
    \begin{itemize}[noitemsep]
        \item Proper names are replaced with other proper names.
        \item Job titles are substituted with similar titles.
        \item Animals are swapped for others with similar traits (\textit{e.g.}, flying animals remain flying).
        \item Important objects retain similar nature and function.
    \end{itemize}
    /* Output Format */
    
    Format your response as a JSON object like this:
    \begin{verbatim}
{
    "character1": ["alternative1", "alternative2"],
    "character2": ["alternative3", "alternative4"],
    "object1": ["alternativeObject1", "alternativeObject2"]
}
    \end{verbatim}
    All names must be in quotation marks. Do not add any extra text or explanation.

    /* Example */
    
    Input: ``The clever Fox and the proud Crow perched on a tall tree.''
    
    Output:
    \begin{verbatim}
{
    "Fox": ["Coyote", "Wolf"],
    "Crow": ["Raven", "Magpie"],
    "tree": ["oak", "pine"]
}
\end{verbatim}
    \end{prompt}
\end{figure*}

\begin{figure*}
\footnotesize
    \begin{prompt}[title={Character Feature Extraction}]
    You are a helpful AI specializing in extracting information from fables and short stories. Analyze the provided fable and identify all characters mentioned in the text. For each character, determine the two most relevant single-word adjectives that best describe their features.
    Format your response as a JSON file, like this:
    \begin{verbatim}
{
    "character1": ["adjective1", "adjective2"],
    "character1": ["adjective3", "adjective4"]
}
    \end{verbatim}
    Make sure to include all relevant characters, regardless of their role in the story, and ensure that the character names are spelled correctly and are in quotation marks. Make sure that each character has two single-word features. Only respond with a json object and nothing else.
    \end{prompt}
\end{figure*}


\begin{figure*}
\footnotesize
    \begin{prompt}[title={Adjective Injection Task}]
    You are an AI that specializes in generating morals for fables and short stories. When given a list of adjectives and a moral, your task is to appropriately inject the adjectives into the moral where they naturally fit, ensuring that the meaning, coherence, and style of the original moral are maintained. The process should include the following steps:
    \begin{itemize}[noitemsep]
        \item Use only the adjectives provided in the list.
        \item Insert the adjectives into the original moral to enhance its description, while keeping the overall length and language style as similar as possible.
    \end{itemize}
    Produce an output formatted strictly as a JSON object with the following keys:
    \begin{itemize}[noitemsep]
        \item "used\_adjectives": a string containing the adjectives you were able to incorporate.
        \item "original\_moral": the unchanged original moral.
        \item "adversarial\_moral": the modified moral with the adjectives injected.
    \end{itemize}
    Return only the JSON object and nothing else. Ensure your response is well-formed JSON.

    /* Example */
    
    Input:
    Adjectives: ['stout', 'mocking', 'weak', 'resilient']
    Moral: Gratitude should be shown through kindness, not harm.

    Eligible Output:
    \begin{verbatim}
{
    "used_adjectives": "stout, resilient",
    "original_moral": "Gratitude should be shown through kindness, not harm.",
    "adversarial_moral": "Stout gratitude should be shown through resilient kindness, not harm."
}
    \end{verbatim}
    \end{prompt}
\end{figure*}


\begin{figure*}
\footnotesize
    \begin{prompt}[title={Adjective-Based Moral Generation}]
    You are an AI specialized in generating morals for fables and short stories. When given a list of adjectives, produce a concise moral that aligns with at least part of the themes indicated by those adjectives. The moral should meet the following criteria:
    \begin{itemize}[noitemsep]
        \item It must be coherent with at least part of the provided adjectives.
        \item It should be written as a short, succinct sentence consisting of only a few words.
        \item The style and length should be similar to the examples provided below.
    \end{itemize}

    Examples:
    \begin{itemize}[noitemsep]
        \item A mild disposition can put a stop to vicious behaviour.
        \item Appearances are deceptive.
        \item In quarreling about the shadow we often lose the substance.
    \end{itemize}

    Output Format:
    Return only a JSON object formatted exactly as follows:
    \begin{verbatim}
{
    "adversarial_moral": ""
}
    \end{verbatim}
    Do not include any additional text or commentary -- only the JSON object should be returned.
    \end{prompt}
\end{figure*}


\begin{figure*}
\footnotesize
    \begin{prompt}[title={Partial Story Moral Generation}]
    /* Instruction */
    You are a creative AI tasked with producing morals for fables and short stories. Your goal is to generate a short moral that is solely based on the partial story you are given, ensuring that the moral is unique, relevant, and not influenced by any external information.

    Instructions:
    \begin{itemize}[noitemsep]
        \item You will be provided with a partial story.
        \item Analyze the specific aspects of the narrative to derive a unique moral lesson.
        \item If the story is very short, enhance it by inventing a small twist or scenario that expands the context, leading to a distinctly new moral.
        \item Ensure that the moral is succinct, fits the narrative's style and tone, and does not refer to outside contexts or general knowledge.
        \item Maintain similar length and language style with the example.
        \item Your final answer must be formatted strictly as a JSON object, with only the JSON present in your response, following the schema below:
    \end{itemize}

    \begin{verbatim}
{
    "moral": "<a moral that conveys a unique lesson>"
}
    \end{verbatim}

    Remember: Base the moral strictly on the partial story and keep the JSON object as your only output.

    /* Example */
    
    INPUT:
    
    Story: ``A Famished Wolf [...] ''

    EXPECTED OUTPUT:
    \begin{verbatim}
{
    "moral": "Fear can lead to unexpected alliances."
}
    \end{verbatim}
    \end{prompt}
\end{figure*}


\begin{figure*}
\footnotesize
    \begin{prompt}[title={Moral Selection Task (MCQA)}]
    You are a helpful AI who specializes in evaluating the moral of fables and short stories.
    Given a fable, you will receive multiple choices of morals, and must select the correct one.
    Only respond with your chosen answer id:
    0, 1, 2, ...

    Here is one example:

    /* Story */
    
    An Ass, carrying a load of wood, passed through a pond. [...]

    /* Choices */
    \begin{enumerate}[label={[\arabic*]}, start=0, noitemsep]
        \item Kindness soothes burdens.
        \item Men often bear little grievances with less courage than they do large misfortunes.
        \item Do not attempt the impossible, lest you become clumsy in your efforts and burdened by the weight of your failure,
        for even the amused onlooker will eventually lose patience with your condescending attitude towards the feat.
        \item Better to endure a small hardship than risk a greater one.
        \item Pride can be a heavy burden, but humility can help you stay afloat in life's challenges.
    \end{enumerate}
/* Expected response */
    \begin{verbatim}
1
    \end{verbatim}
    \end{prompt}
\end{figure*}


\begin{figure*}
\footnotesize
    \begin{prompt}[title={Moral Selection Task (Binary opposites)}]
    You are a helpful AI who specializes in evaluating the moral of fables and short stories.
    Given a fable, you will receive 2 choices of morals, and must select the correct one. Only respond with the answer id (0 or 1) and nothing else.
If the fable is incomplete or missing, use your best judgment to determine the most appropriate moral.
    Here is one example:

    /* Story */
    
   An Ant nimbly running about in the sunshine in search of food came across a Chrysalis [...]

    /* Choices */
  \begin{enumerate}[label={[\arabic*]}, start=0, noitemsep]
        \item Appearances are deceptive.
        \item Appearances are truthful.
  \end{enumerate}

/* Expected response */
    \begin{verbatim}
0
    \end{verbatim}

    \end{prompt}
\end{figure*}


\begin{figure*}
\footnotesize
    \begin{prompt}[title={Moral Selection Task (True or False evaluation)}]
    You are a helpful AI who specializes in evaluating the moral of fables and short stories.
    For each fable provided, you will receive a proposed moral. Your task is to determine whether the moral accurately reflects the fable. Respond only with \textit{True} or \textit{False}.
 Here are two examples for the same story:

    Example 1:
    
    /* Story */
    
    An Ass, carrying a load of wood [...]

    /* Question */
    
    True or False: The moral is: ``Men often bear little grievances with less courage than they do large misfortunes.''

    /* Expected response */
    \begin{verbatim}
True
    \end{verbatim}

    Example 2:
    
    /* Story */
    
    An Ass, carrying a load of wood [...]

    /* Question */
    
    True or False: The moral is: ``Watch the actions of your enemy.''

    /* Expected response */
    \begin{verbatim}
False
    \end{verbatim}
    \end{prompt}
\end{figure*}


\begin{figure*}
\footnotesize
    \begin{prompt}[title={Story Adjective Injection Task}]
    You are a helpful AI specializing in producing fables and short stories. When provided with a list of adjectives and a story, your task is to enrich the story by injecting the provided adjectives into it, but only if they naturally fit and enhance the meaning of the text. You must integrate the adjectives with minimal modification to the original text, avoiding the creation of new sentences or the completion of incomplete ones.

    Instructions:
    \begin{itemize}[noitemsep]
        \item Analyze the provided story for appropriate opportunities where adjectives from the list can be seamlessly added.
        \item Modify the original story only as much as needed to incorporate the adjectives, ensuring that the result remains coherent and retains the original story's length and style.
        \item Only use adjectives provided in the input list. Do not substitute, omit, or add adjectives from outside the supplied list.
        \item If the story text ends in an incomplete sentence or thought, do not attempt to complete it; simply inject adjectives where possible, leaving abrupt endings intact.
    \end{itemize}

    Output Format:
    Return your response as a JSON object exactly like the following structure, without any extra commentary or modifications:
    \begin{verbatim}
{
    "used_adjectives": "<comma-separated list of adjectives that were added>",
    "new_story": "<the original story with the suitable adjectives injected>"
}
    \end{verbatim}

    Ensure that your JSON is valid and that the keys and content strictly follow the structure defined above.
    \end{prompt}
\end{figure*}


\begin{figure*}
\footnotesize
    \begin{prompt}[title={Moral Extraction Task}]
    You are a helpful AI that specializes in analyzing the morals of fables and short stories. When provided with a story, your task is to determine and output its moral. Use the style, tone and length shown in the examples, following a traditional Aesopian style.

    /* Guidelines */
    \begin{itemize}[noitemsep]
        \item Read the provided story carefully.
        \item Identify the underlying moral or lesson embedded in the narrative.
        \item Use clear and concise language similar to the examples.
        \item Do not include any extraneous notes, commentary, or explanations—output only the final moral.
    \end{itemize}

    /* Output Format */
    
    Your entire output should be the moral statement, with no additional text, with no quotation marks.

    /* Examples */
    
    Example 1:
    
    Story:
    
    ``One summer's day a Grasshopper was hopping about, chirping and singing to its heart's content. [...]''

    Expected response: 
    
    There's a time for work and a time for play.

    Example 2:
    
    Story: ``There was a groom who used to sell his horse's barley to the innkeepers and drink all evening long. [...] ''

    Expected response: 
    
    Someone who wants to help his friend must give him what is essential and appropriate.

    Example 3:
    
    Story: ``A reed got into an argument with an oak tree.  [...]''

    Expected response: 
    
    Those who adapt to the times will emerge unscathed.

    \end{prompt}
\end{figure*}


\end{document}